\documentclass{article} 
\usepackage{iclr2025_conference,times}
\usepackage[hyperindex,breaklinks,colorlinks,citecolor=darkblue]{hyperref}

\usepackage{amsmath,amsfonts,bm}

\def\eqref#1{equation~\ref{#1}}

\def\1{\bm{1}}

\def\mC{{\bm{C}}}

\def\mF{{\bm{F}}}

\def\mI{{\bm{I}}}

\def\mR{{\bm{R}}}

\def\mY{{\bm{Y}}}

\DeclareMathAlphabet{\mathsfit}{\encodingdefault}{\sfdefault}{m}{sl}
\SetMathAlphabet{\mathsfit}{bold}{\encodingdefault}{\sfdefault}{bx}{n}

\usepackage[textwidth=3cm]{todonotes}

\usepackage{hyperref}
\usepackage{url}
\usepackage{xspace}
\usepackage{graphicx}
\usepackage{enumitem}
\usepackage{wrapfig}
\usepackage{subfigure}
\usepackage{booktabs}
\usepackage{tabularx}
\usepackage{multirow}
\usepackage{transparent}
\usepackage{caption}
\usepackage[toc, page, header]{appendix}
\usepackage[capitalize,noabbrev]{cleveref}
\newcommand{\alg}{TRACE\xspace}
\newcommand{\framework}{causal event modeling\xspace}

\title{\alg: Temporal Grounding Video LLM  via Causal Event Modeling}

\author{Yongxin Guo$^{1}${\quad\,}
Jingyu Liu$^{2}${\quad\,}
Mingda Li$^{2}${\quad\,}
Qingbin Liu$^{2}${\quad\,}
Xi Chen$^{2,}$$^*${\quad\,}
Xiaoying Tang$^{1,3,4,}$\thanks{Corresponding authors.}
\\
$^1$School of Science and Engineering, The Chinese University of Hong Kong, Shenzhen 518172, China\\
$^2$Tencent PCG \\
$^3$Shenzhen Institute of Artificial Intelligence and Robotics for Society (AIRS), Shenzhen, China\\
$^4$Guangdong Provincial Key Laboratory of Future Networks of Intelligence, Shenzhen, China\\
}

\iclrfinalcopy 
\begin{document}

\definecolor{darkblue}{rgb}{0.0, 0.0, 0.55}

\maketitle

\begin{abstract}
Video Temporal Grounding (VTG) is a crucial capability for video understanding models and plays a vital role in downstream tasks such as video browsing and editing. 
To effectively handle various tasks simultaneously and enable zero-shot prediction, there is a growing trend in employing video LLMs for VTG tasks. However, current video LLM-based methods rely exclusively on natural language generation, lacking the ability to model the clear structure inherent in videos, which restricts their effectiveness in tackling VTG tasks. To address this issue, this paper first formally introduces \framework framework, which represents {video LLM outputs} as sequences of events, and predict the current event using previous events, video inputs, and textural instructions. Each event consists of three components: timestamps, salient scores, and textual captions. We then propose a novel task-interleaved video LLM called \alg to effectively implement the \framework framework in practice. 
The \alg {process} visual frames, timestamps, salient scores, and text as distinct tasks, employing various encoders and decoding heads for each. Task tokens are arranged in an interleaved sequence according to the \framework framework's formulation.
Extensive experiments on various VTG tasks and datasets demonstrate the superior performance of \alg compared to state-of-the-art video LLMs. Our model and code are avaliable at \url{https://github.com/gyxxyg/TRACE}.
\end{abstract}

\section{Introduction}

Video Temporal Grounding (VTG) is an important ability for video understanding models~\citep{lin2023univtg}, and has becoming the base of a series of downstream tasks like moment retrieval~\citep{caba2015activitynet,gao2017tall,oncescu2021queryd}, dense video caption~\citep{zhou2018towards,tang2019coin}, video highlight detection~\citep{lei2107qvhighlights,liu2022umt}, and video summarization~\citep{song2015tvsum,gygli2014creating}.
While non-generative models excel in moment retrieval and video highlight detection~\citep{lei2021detecting,han2024unleash,wang2024internvideo2}, they are inflexible, task-specific, and demand substantial fine-tuning for optimal performance. 
To tackle these challenges, recent research employs video LLMs as versatile models, integrating timestamp information into visual inputs, and fine-tuning them on VTG tasks~\citep{ren2023timechat,huang2023vtimellm,wang2024hawkeye,qian2024momentor,wang2024efficient,wu2024number} to enhance their performance and facilitate zero-shot prediction.

\begin{figure}[!t]
    \centering

    \subfigure[Video Structure]{
    \includegraphics[width=.45\linewidth]{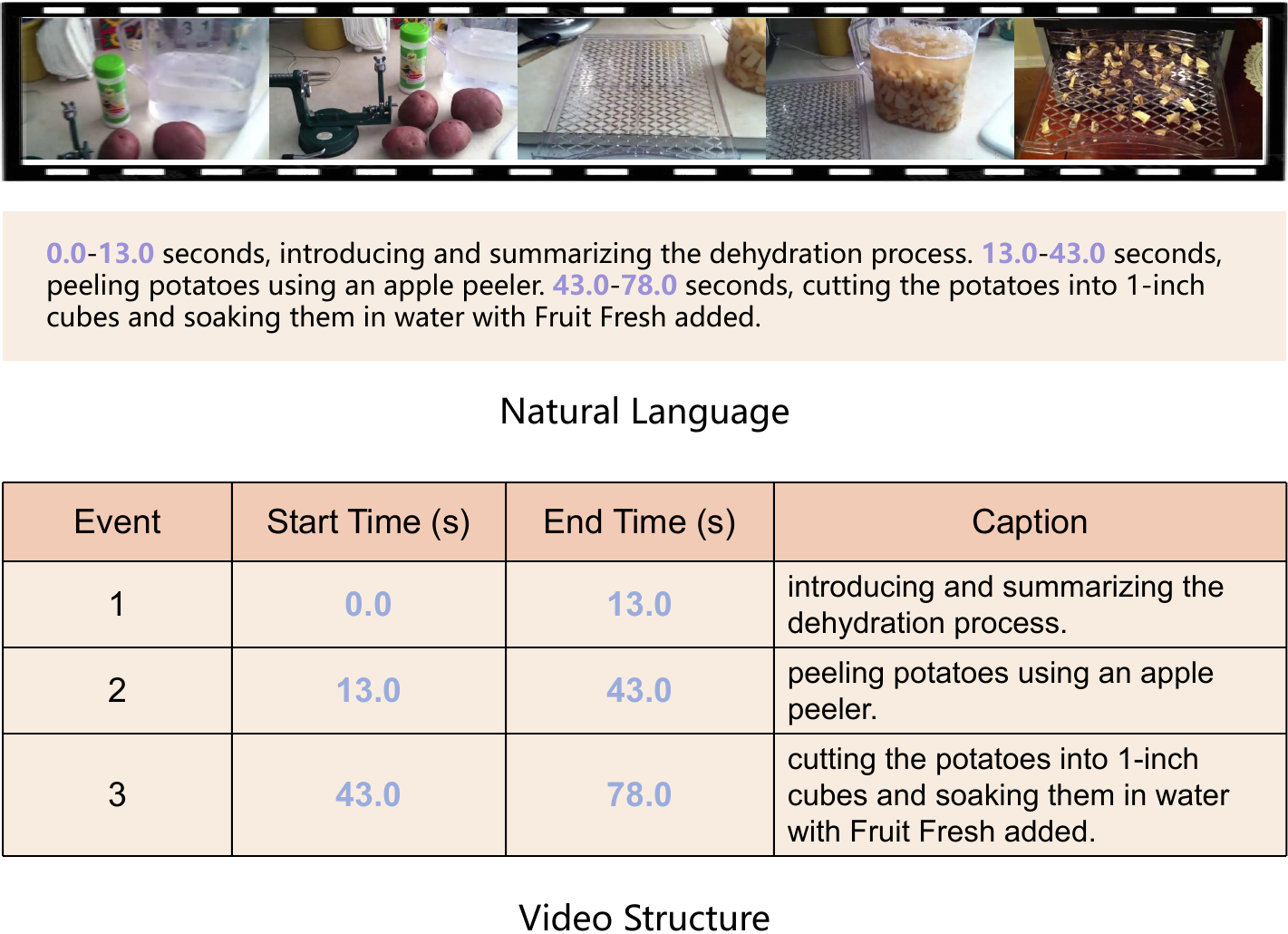}
    \label{fig:video-structure-illustration}
    }
    \subfigure[Performance Gap between Models]{
    \includegraphics[width=.45\linewidth]{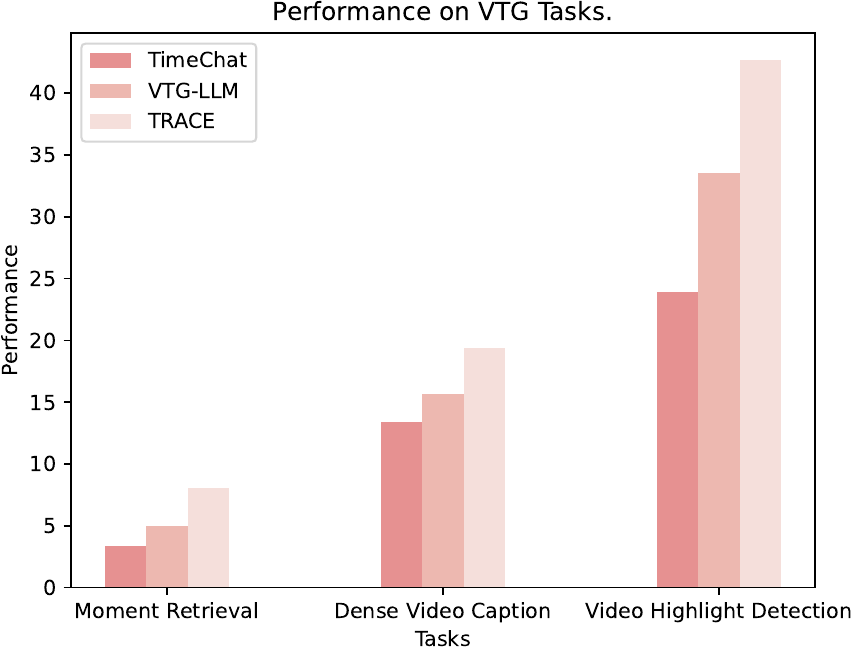}
    \label{fig:challenge-illustrate}
    }
    \caption{\small \textbf{Challenges posed by videos' inherent structures.} Figure~\ref{fig:video-structure-illustration} shows the difference between natural language and video structure, while Figure~\ref{fig:challenge-illustrate} highlights the performance gap between SOTA video LLMs~\citep{ren2023timechat,guo2024vtg} and \alg. We present zero-shot performance results for video LLM approaches. Specifically, we report the performance of models using the CIDEr metric for the dense video captioning task on the Youcook2 dataset, R@1$_{\text{IOU=0.7}}$ for the moment retrieval task on the Charades-STA dataset, and HIT@1 for the highlight detection task on the QVHighlights dataset.\looseness=-1}
    \vspace{-1.5em}
\end{figure}

\textbf{Challenges posed by videos' inherent structures.}
Despite reflecting human intent, current video LLM based approaches rely on pure natural language generation. {As illustrated in Figure~\ref{fig:video-structure-illustration}, this approach lacks a clear structure and indiscriminately blends information, such as timestamps and text captions.}
{In contrast, videos possess an inherent structure that transcends mere textual description.  To accurately describe or reason from a video, it is insufficient to rely solely on natural language text. Instead, the corresponding timestamps and salient scores are also essential components. Together, these elements provide a more comprehensive and structured understanding of the video content.}
Consequently, the gap between {videos' structure} and current video LLMs undermines the ability of video LLMs to effectively model video events, potentially making video LLMs difficult to achieve satisfactory results (Figure~\ref{fig:challenge-illustrate}) on VTG tasks.

\begin{figure}[!t]
    \centering
    \includegraphics[width=1.\linewidth]{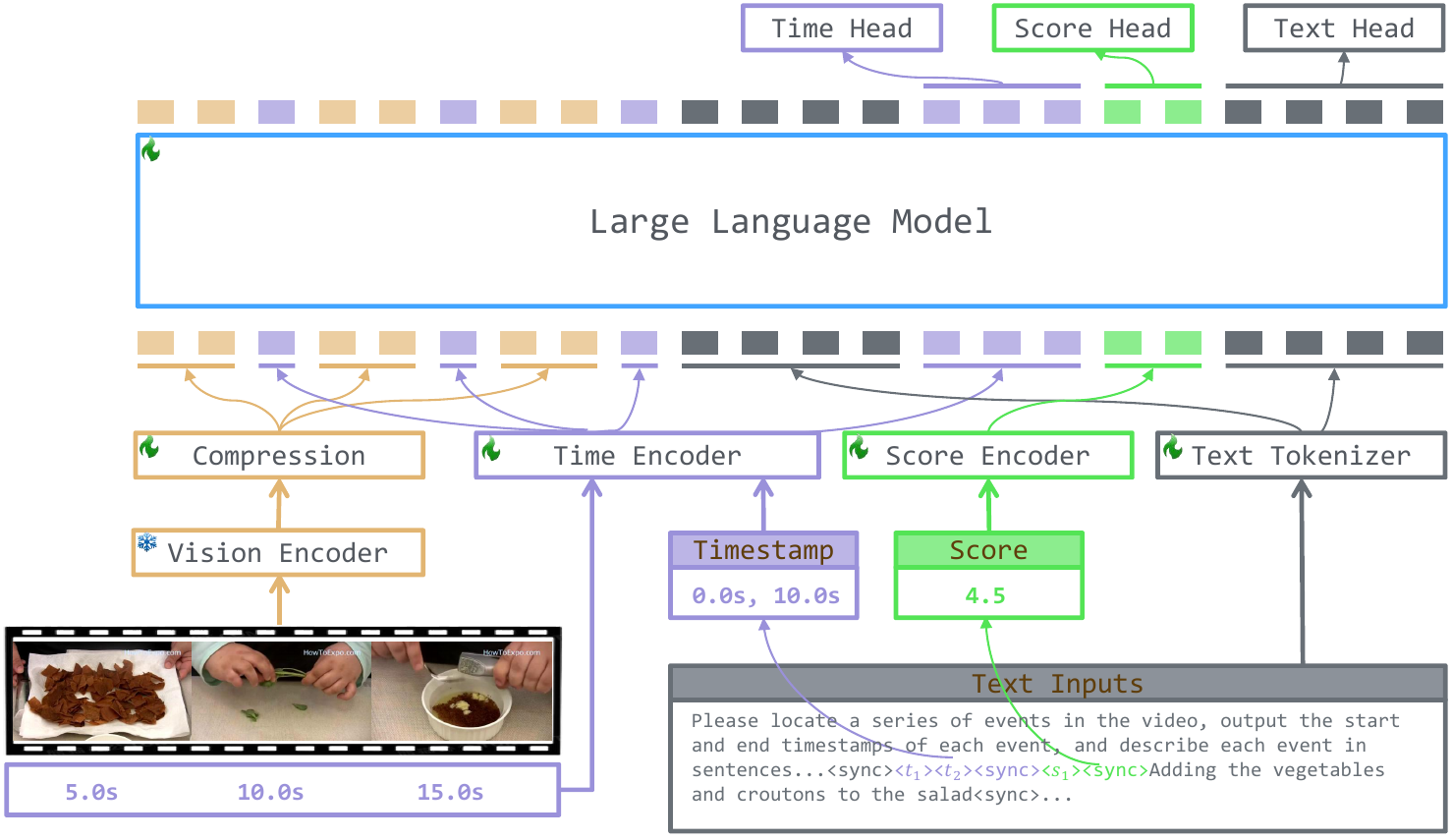}
    \caption{\small \textbf{Overview of the training process of \alg model.} We employ a variety of encoders and heads to handle \textcolor[rgb]{0.6,0.57,0.85}{time}, \textcolor[rgb]{0.55,0.93,0.56}{score}, and \textcolor[rgb]{0.41,0.44,0.46}{text} inputs and outputs. The timestamps of the sampled frames are converted into time tokens and subsequently integrated into the visual tokens of each frame. In the answer section, time tokens, score tokens, and text tokens are inserted in a sequential manner. The generation process of \alg is summarized in Figure~\ref{fig:generation-illustration}.}
    \label{fig:nero-overview}
\end{figure}

\textbf{Causal event modeling as a solution.} 
In this paper, our primary goal is to develop a novel video LLM approach for resolving the mismatch between language modeling of LLMs and videos' inherent structure. 
Specifically, we concentrate on tackling two main challenges: (1) developing a theoretical framework that shifts from causal language modeling to structured {event-based} modeling, and (2) constructing a practical video LLM based on the theoretical framework to provide an effective solution.
To accomplish this, we first introduce the \framework framework, where {video LLM outputs} are represented as sequences of events, each containing timestamps, salient scores, and textual captions. The next events are predicted based on video inputs, text instructions, and preceding events.
To effectively implement the \framework framework in practice, we present a novel task-interleaved video LLM, TempoRAl grounding via Causal Event modeling (\alg), as illustrated in Figure~\ref{fig:nero-overview}.
The \alg treats visual frames, timestamps, salient scores, and text as separate tasks, utilizing diverse encoders and decoding heads for each task, with task tokens sequenced in an interleaved manner.
Furthermore, we develop an adaptive head-switching method for improved generation. Our numerical results across various VTG tasks reveal the superior performance of \alg in comparison to state-of-the-art (SOTA) video LLMs.

\textbf{Our key contributions are summarized as follows:}
\begin{itemize}[nosep,leftmargin=12pt]
    \item We model the videos by a series of events, and propose \framework framework to capture videos' inherent structure. We then present a novel task-interleaved video LLM model, \alg, tailored to implement the \framework framework through the sequential encoding/decoding of timestamps, salient scores, and textual captions.
    \item We conduct comprehensive experiments on multiple VTG tasks and datasets to verify the effectiveness of \alg. The results reveal significant improvements of \alg in comparison to SOTA video LLMs. Notably, \alg improves zero-shot performance by 3.1 and 4.9\% on Youcook2 (CIDEr and F1 Score), by 6.5\% and 3.7\% in Recall (IOU = \{0.5, 0.7\}) on Charades-STA, and by 10.3\% and 9.2\% for mAP and HIT@1 on QVHighlights. Moreover, surpassing existing video LLMs, \alg achieves comparable performance to traditional non-generative and task-specific methods after fine-tuning, highlighting the potential of video LLMs to excel in VTG tasks.
\end{itemize}


\section{Related Works}

\paragraph{Video temporal grounding.} Video Temporal Grounding (VTG) tasks aim to precisely identify the timestamps of events within a given video~\citep{lin2023univtg}. This includes various tasks such as moment retrieval~\citep{gao2017tall,Zala2023HiREST,oncescu2021queryd,hendricks18emnlp,boris2024surprising}, dense video caption~\citep{zellersluhessel2021merlot,Zala2023HiREST,tang2019coin,caba2015activitynet,kim2024you}, video summarization~\citep{song2015tvsum,gygli2014creating,hua2024v2xum}, and video highlight detection~\citep{lei2021detecting,xiao2023bridging}. 
For tasks such as moment retrieval, video summarization, and video highlight detection, traditional approaches primarily use large-scale video-text pre-training~\citep{xu2021videoclip,wang2022internvideo,yan2022videococa,li2023unmasked,chen2024vast,tong2022videomae,zhao2024videoprism}. Subsequently, they fine-tune the pretrained models by incorporating task-specific prediction heads.
While these methods have demonstrated satisfactory results, they are resource-intensive for pre-training, lack zero-shot capabilities, can only handle one specific task per model, and often require additional fine-tuning for numerous downstream tasks.
For the dense video caption task, Vid2Seq employs special time tokens to represent timestamps~\citep{yang2023vid2seq}. Some approaches integrate additional input information, such as text queries from training datasets~\citep{kim2024you}, while other models utilize different decoding heads to decode timestamps and textual captions~\citep{wang2021end,wang2023learning} in parallel. However, these architectures are specifically designed for the dense video caption task, cannot be easily adapted to LLM structures to fully harness the capacity of pretrained LLMs, and also lack zero-shot capabilities.

\paragraph{Video LLMs for video temporal grounding.} Large language models (LLMs)~\citep{kaplan2020scaling,achiam2023gpt,touvron2023llama} have demonstrated significant potential in acquiring knowledge and addressing real-world challenges using a zero-shot approach. Recent research has focused on integrating knowledge from other modalities, such as vision~\citep{liu2024visual,li2023blip} and audio~\citep{ghosal2023text}, to bolster the capabilities of LLMs. Within the visual domain, video large language models (video LLMs) have emerged as a crucial research area~\citep{lin2023video,maaz2023video, zhu2023minigpt,song2024moviechat,song2024moviechat+}.
Traditional video LLMs~\citep{zhang2023video, lin2023video, li2023videochat,li2024llava,cheng2024videollama,yao2024minicpm} have made considerable performance improvements in tasks such as video question answering, reasoning, and video captioning. However, these methods encounter difficulties in precisely pinpointing event timestamps within videos. To address this issue, TimeChat~\citep{ren2023timechat}, VTimeLLM~\citep{huang2023vtimellm}, and Hawkeye~\citep{wang2024hawkeye} have attempted to overcome this limitation by fine-tuning the video LLMs on VTG datasets.
More recently, LITA~\citep{huang2024lita} introduces fast-slow visual tokens and incorporates time tokens into LLM tokenizers. Momentor~\citep{qian2024momentor} suggests a time encoder to address time token quantization errors.
VTG-LLM~\citep{guo2024vtg} integrates special time tokens and time position embeddings to improve the ability of video LLMs in comprehending timestamps.
However, these methods do not take into account the inherent structure of videos and still cannot achieve satisfactory performance. In this paper, we propose the \framework framework to {provide structured video LLM outputs} and design the \alg model to address the proposed framework. Numerical results demonstrate significant performance gains of \alg over existing video LLMs on VTG tasks.
\section{\alg}

In this section, we aim to develop a novel video LLM that aligns well with video structures, addressing two questions: (1) how to model {the structured video LLM outputs that are align well with video structures}, and (2) how to implement theoretical models. We start by proposing \emph{\framework} framework to tackle "how to model". Then, we introduce \alg to address "how to implement".
We have included a detailed discussion about our framework in Appendix~\ref{sec:discussion}.

\subsection{Modeling the Inherent Structures of Videos}

\paragraph{Formulating {outputs of video LLMs} by events.} 

{
Given the instruction $\mathbf{I}$ and video visual inputs $\mathbf{F}$, we represent the outputs of video LLMs $\mathbf{R}$ as a series of events $\{ e_1, e_2, \cdots, e_K \}$, with each event $e_k = (t_k, s_k, c_k)$ encompassing timestamps $t_k$, salient scores $s_k$, and textual captions $c_k$. In summary, we have
\begin{align}
    \mathbf{R} = \{ e_1, e_2, \cdots, e_K \} = \{ (t_k, s_k, c_k) | 1 \le k \le K \} \, .
\end{align}
}
\paragraph{Causal event modeling framework.} To effectively utilize the knowledge of pretrained LLMs, the design of \framework shares the underlying intuition of causal language modeling, as formulated in the subsequent equation~\footnote{Theoretically, the order of time, score, and text will not impact the results. We select one order here.}.
\begin{align}
    \mathcal{P}(e_k | e_{1:k-1}, \mathbf{I}, \mathbf{F}) & = 
    \mathcal{P}(t_k, s_k, c_k | e_{1:k-1}, \mathbf{I}, \mathbf{F}) \, , \nonumber \\
    & = \mathcal{P}(t_k | e_{1:k-1}, \mathbf{I}, \mathbf{F})
    \mathcal{P}(s_k | t_k, e_{1:k-1}, \mathbf{I}, \mathbf{F}) \mathcal{P}(c_k | s_k, t_k, e_{1:k-1}, \mathbf{I}, \mathbf{F}) \, ,
    \label{equ:next-event-prediction}
\end{align}
The next event $e_k$ is determined by textural instructions, visual inputs, and previous events. 
We can find that \framework framework aligns well with the video structure (Figure~\ref{fig:video-structure-illustration}):
(1) timestamps, salient scores, and textual captions are sequentially decoded within each event;
(2) events are then ordered by timestamps.


\subsection{\alg: Task-Interleaved Temporal Grounding Video LLM}

In Eq.~\ref{equ:next-event-prediction}, we introduce a formal \framework framework to tackle the challenge of modeling {structured video LLM outputs}. This section illustrates the design of \alg to implement the \framework framework (Figure~\ref{fig:nero-overview}).

\paragraph{Overview of \alg.} As illustrated in Eq.~\ref{equ:next-event-prediction}, the \framework framework necessitates encoding/decoding of visual frames ($\mathbf{F}$), text ($\mathbf{I}$ and $c_k$), timestamps ($t_k$), and scores ($s_k$). Consequently, the \alg considers these elements as distinct tasks and employs the following design to efficiently manage them.
\begin{itemize}[nosep,leftmargin=12pt]
    \item \textit{Separated multi-task processing.} \alg utilizes separate encoders and decoding heads for each task to convert task inputs into task tokens and decode task tokens back to outputs (Sec.~\ref{sec:task-processing}). 
    \item \textit{Task-interleaved sequence modeling.} Task tokens are sequenced in an interleaved manner according to Eq.~\ref{equ:next-event-prediction} in \alg and fed into LLM backbones (Sec.~\ref{sec:sequence-modeling}).
    \item \textit{Adaptive head-switching mechanism for generation.} During generation, we implement an adaptive head-switching mechanism to select the appropriate decoding head for producing the next token (Sec.~\ref{sec:generation}).
\end{itemize}

\subsubsection{Separated Multi-Task Processing}
\label{sec:task-processing}

\alg consists of four unique tasks: visual frames, text, timestamps, and scores. Regarding text, we directly utilize the text tokenizer and LLM head of the LLM backbone (Mistral-7B-v0.2~\citep{jiang2023mistral}). 
Moreover, we added a special token $\langle sync \rangle$ for indicating the end of text tasks.
The processing for the other tasks is detailed below.

\paragraph{Timestamps and scores processing.} For processing timestamps and score information, we employ two separate encoders and decoding heads, both of which share the same architecture. Specifically, each encoder is initialized with a tokenizer containing 13 tokens: 11 number tokens $\langle 0 \rangle, \cdots, \langle 9 \rangle, \langle . \rangle$ for representing timestamps/scores, $\langle sep \rangle$ to mark the end of each timestamp/score, and $\langle sync \rangle$ to signify the end of the current task. Token embeddings are initialized using LLM token embeddings.

In accordance with the research in VTG-LLM~\citep{guo2024vtg}, we format each timestamp/score to the same length, comprising 4 whole-number parts, 1 dot, and 1 fractional part~\footnote{Different from timestamps, scores will be encoded to 3 score tokens, including 1 whole-number parts, 1 dot, and 1 fractional part.}. Subsequently, $\langle sep \rangle$ is inserted between timestamps/scores, and $\langle sync \rangle$ is appended at the end of each timestamp/score input sequence. For instance, the timestamp inputs [10.23, 125.37] will be tokenized into the following sequence: $\langle 0 \rangle \langle 0 \rangle \langle 1 \rangle \langle 0 \rangle \langle . \rangle \langle 2 \rangle \langle sep \rangle \langle 0 \rangle \langle 1 \rangle \langle 2 \rangle \langle 5 \rangle \langle . \rangle \langle 4 \rangle \langle sync \rangle$.

\paragraph{Visual frames processing.} Given a $T$-frame video, we initially encode the frames using the pretrained CLIP ViT-L~\citep{radford2021learning}, with each frame being encoded into 576 visual tokens. Subsequently, we employ Slot-Based Compression~\citep{guo2024vtg} to reduce the number of visual tokens to 8 per frame.
Moreover, to integrate temporal information into the visual inputs, we use a time encoder to encode the timestamps of each sampled frame and remove the $\langle sync \rangle$ and $\langle sep \rangle$ tokens, resulting in 6 time tokens for each frame. Finally, we concatenate the 8 visual tokens with the 6 time tokens to form the visual inputs for each frame.

\begin{figure}[!t]
    \centering
    \includegraphics[width=1.\linewidth]{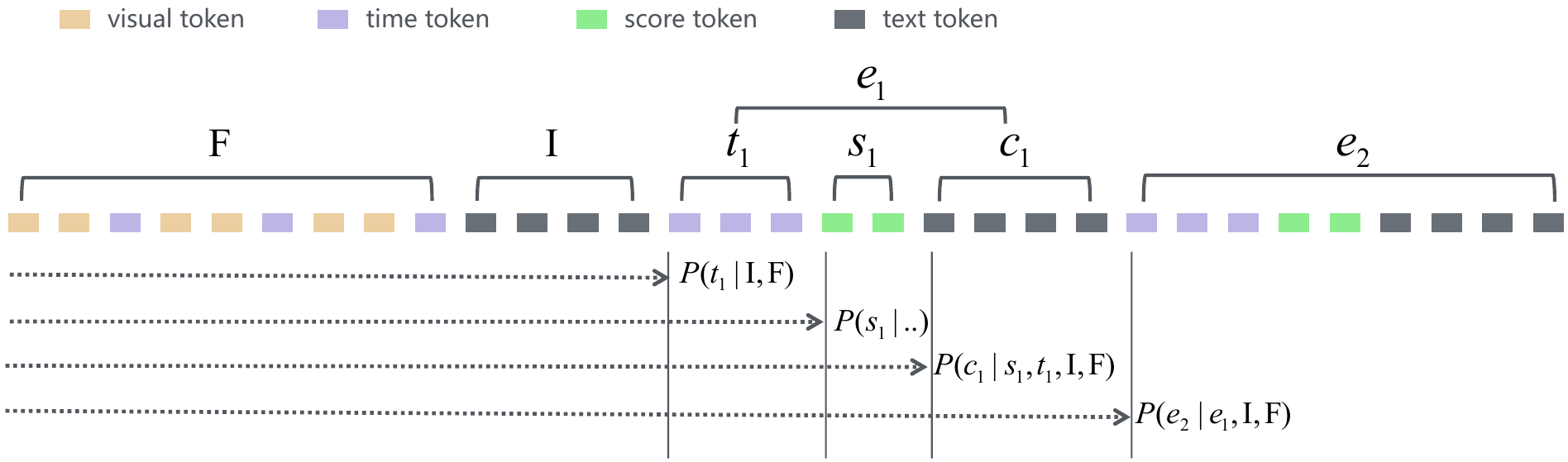}
    \caption{\small \textbf{Illustration of token sequence of \alg.} Following Eq~\ref{equ:next-event-prediction}, the sequence begins with visual frame tokens ($\mathbf{F}$) followed by instruction tokens ($\mathbf{I}$). Event ($e$) tokens are structured in the sequence of time tokens ($t$), score tokens $s$, and text tokens $c$, with events ordered chronologically based on their occurrence time.}
    \vspace{-1em}
    \label{fig:sequence-illustration}
\end{figure}

\subsubsection{Task-interleaved sequence modeling}
\label{sec:sequence-modeling}

Utilizing the processed task tokens, we construct the sequence following Eq.~\ref{equ:next-event-prediction}. The token sequence order is illustrated in Figure~\ref{fig:sequence-illustration}. 
\paragraph{Inter-event sequence order.} The sequence commences with visual frame tokens $\mathbf{F}$ followed by textual instruction tokens $\mathbf{I}$. For the events section, event tokens are sequenced according to the events' occurrence time to align with the \framework formula $\mathcal{P}(e_k | e_{1:k-1}, \mathbf{I}, \mathbf{F})$.

\paragraph{Intra-event sequence order.} For each event, in accordance with Eq.~\ref{equ:next-event-prediction}, tokens are arranged sequentially by time tokens ($\mathcal{P}(t_k | e_{1:k-1}, \mathbf{I}, \mathbf{F})$), score tokens ($\mathcal{P}(s_k | t_k, e_{1:k-1}, \mathbf{I}, \mathbf{F})$), and text tokens ($\mathcal{P}(c_k | s_k, t_k, e_{1:k-1}, \mathbf{I}, \mathbf{F})$). Consequently, the \framework framework (Eq.~\ref{equ:next-event-prediction}) emerges as a specialized autoaggressive model, featuring a unique sequence order that closely aligns with video structures.

\subsubsection{Adaptive Head-Switching Mechanism for Generation}
\label{sec:generation}

\begin{figure}[!t]
    \centering
    \includegraphics[width=1.\linewidth]{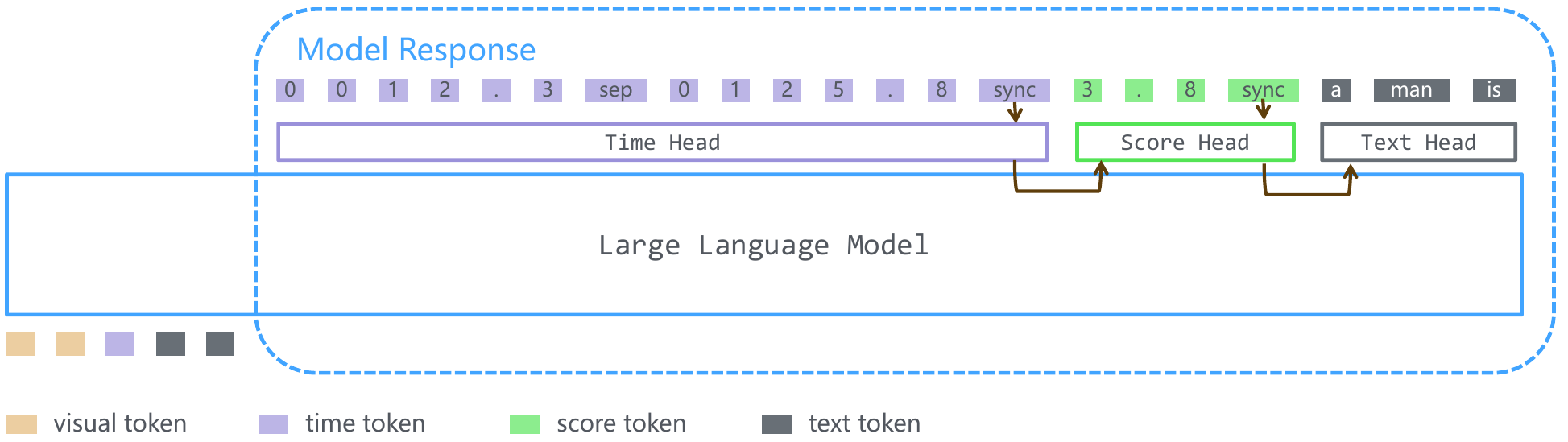}
    \caption{\small \textbf{Generation process of \alg.} The \alg generate tokens following the order of time tokens, score tokens, and text tokens. The decoding heads are switched when $\langle sync \rangle$ tokens are generated.}
    \vspace{-1em}
    \label{fig:generation-illustration}
\end{figure}

\paragraph{Using $\langle sync \rangle$ token for adaptive head switching.} Since \alg employs distinct decoding heads for various tasks during training, selecting the appropriate decoding head during generation based on previously decoded tokens is crucial. This selection is facilitated by the $\langle sync \rangle$ token. As illustrated in Figure~\ref{fig:generation-illustration}, \alg generates tokens in the sequence of time, score, and text tokens. Detection of the $\langle sync \rangle$ token prompts \alg to switch decoding heads accordingly. The heads are cycled switched in the order of time head - score head - text head.

\subsection{Training Strategy and Data Preparation}
\label{sec:train-process}

This section outlines the \alg training process, which includes two stages.
For the stage 1, task modules such as the vision compression layer, task encoder, and task heads are trained for initialization. For the stage 2, the LLM backbone is fine-tuned while keeping the task modules tuned. Detailed settings and datasets are presented below.
Due to the page limitation, detailed annotation examples for each task, and details about data filtering and processing are provided in Appendix~\ref{sec:data-prepare}.

\paragraph{Stage 1: Initialization of task modules.} In stage 1, task modules such as the vision compression layer, time encoder/head, score encoder/head, and text tokenizer/head are trained while the vision encoder and LLM backbone remain fixed. 
As shown in Table~\ref{tab:datasets}, stage 1 primarily utilizes two groups of datasets.
\begin{itemize}[nosep,leftmargin=12pt]
    \item \textit{Image and video caption datasets for initializing the visual compression layer.} This group of datasets including Valley~\citep{luo2023valley}, LLaVA\_Image~\citep{liu2024visual}, TextVR~\citep{wu2025large}, and a randomly sampled subset of ShareGPT4Video~\citep{chen2024sharegpt4video} datasets.
    \item \textit{VTG datasets for task encoder/head initialization.} We use VTG-IT dataset in this group.
\end{itemize}
For stage 1 training, we uniformly sample 128 frames from each video. The learning rate is set to 1e-3, and models are trained for one epoch with a batch size of 128. 

\paragraph{Stage 2: Instruction tuning for enhancing VTG capacity.} In Stage 2, the LLM backbone and task modules are fine-tuned, with only the vision encoder remaining fixed. 
As shown in Table~\ref{tab:datasets}, stage 2 primarily utilizes three groups of datasets.
\begin{itemize}[nosep,leftmargin=12pt]
    \item \textit{VTG instruction tuning datasets for enhancing VTG capacity.} We use VTG-IT~\citep{guo2024vtg}, ActivityNet Captions~\citep{caba2015activitynet}, and a subset of InternVid~\citep{wang2023internvid}, resulting in a total of 635K data samples. Low-quality samples were filtered out, and the VTG-IT-VHD and VTG-IT-VS datasets were re-annotated. Additional details can be found in Appendix~\ref{sec:data-prepare}.
    \item \textit{Video caption datasets for maintaining the quality of the visual compression layers.} We use parts of the video data from stage 1, such as Valley~\citep{luo2023valley}, TextVR~\citep{wu2025large}, and ShareGPT4Video~\citep{chen2024sharegpt4video} datasets. These datasets are compressed by retaining only one sample for samples with identical videos but different instructions, yielding 284K data.
    \item \textit{Video question answering datasets to enhance \alg's reasoning capabilities.} We use VideoChatGPT~\citep{maaz2023video} and Next-QA~\citep{xiao2021next} in this part.
\end{itemize}
For each video, the content is uniformly divided into 128 clips, with one frame randomly sampled from each clip. The learning rate is set to 5e-6, and the models are trained for two epochs using a batch size of 128.

\newcolumntype{Y}{>{\arraybackslash}m{0.1\textwidth}}
\newcolumntype{Z}{>{\arraybackslash}m{0.7\textwidth}}

\begin{table}[!t]
    \centering
    \caption{\small \textbf{Datasets used for \alg training process.} "Compressed" indicates that datasets are condensed by retaining only one sample for samples with identical videos but varying instructions.}
    \begin{tabularx}{\textwidth}{
    Y Z Y
    }
    \toprule
    Stage & Datasets & Quantity \\
    \midrule
    Stage 1 & Valley, LLaVA\_Image, TextVR, ShareGPT4Video, VTG-IT & 1.9M \\
    \midrule
    Stage 2 & Valley (Compressed), TextVR (Compressed), ShareGPT4Video (Compressed), VTG-IT, ActivityNet Captions, VideoChatGPT, InternVid, Next-QA  & 0.9M \\
    \bottomrule
    \end{tabularx}
    \label{tab:datasets}
    \vspace{-1em}
\end{table}

\section{Experiments}

Detailed experimental settings and hyper-parameters can be found in Appendix~\ref{sec:experiment-setting}. 
Numerical results on more video understanding benchmarks and more ablation studies can be found in Appendix~\ref{sec:additional-experiments}.
Case studies can be found in Appendix~\ref{sec:case-study}.

\subsection{Evaluation Datasets, Metrics, and Baseline Models.}

We evaluate the model performance on three different tasks:
\begin{itemize}[nosep,leftmargin=12pt]
    \item \textit{Dense video caption.} 
    We use Youcook2~\citep{zhou2018towards} and ActivityNet Captions~\citep{caba2015activitynet} datasets as the evaluation datasets. The evaluation metrics include CIDEr~\citep{vedantam2015cider}, METEOR~\citep{banerjee2005meteor}, and SODA\_c~\citep{fujita2020soda} for assessing the quality of the captions. These metrics are averaged under different IoU thresholds $\{0.3, 0.5, 0.7, 0.9 \}$, following previous studies~\citep{ren2023timechat,huang2023vtimellm}. Additionally, we report the F1 score to measure the model's ability to accurately locate timestamps.
    \item \textit{Moment retrieval.}
    We utilize test set of Charades-STA~\citep{gao2017tall} for the moment retrieval task and report the recall at IOU thresholds of 0.5 and 0.7. Additionally, we present the mIOU results.
    \item \textit{Video highlight detection.} We employ the validation set of the QVHighlights dataset~\citep{lei2021detecting} and report the mean average precision (mAP) with IOU thresholds of 0.5 and 0.75, as well as the HIT@1, which represents the hit ratio of the highest scored clip.\looseness=-1
\end{itemize}
For baseline models, we select Valley~\citep{luo2023valley}, VideoChat~\citep{li2023videochat}, Video-ChatGPT~\citep{maaz2023video}, and Video-LLaMA~\citep{zhang2023video} as examples of traditional video LLMs. For video LLMs specifically designed for VTG tasks, we choose TimeChat~\citep{ren2023timechat}, VTimeLLM~\citep{huang2023vtimellm}, Momentor~\citep{qian2024momentor}, HawkEye~\citep{wang2024hawkeye}, and VTG-LLM~\citep{guo2024vtg}.

\subsection{Performance of \alg}

\paragraph{Superior zero-shot performance of \alg over other video LLMs.} In Table~\ref{tab:zero-shot-vtg}, we show the zero-shot performance of \alg compare to SOTA video LLM baselines. 
The results show that
\begin{itemize}[leftmargin=12pt,nosep]
    \item \textbf{Suprior zero-shot performance.} As shown in Table~\ref{tab:zero-shot-vtg}, \alg significantly outperforms other video LLMs by a substantial margin across all three datasets. Notably, it achieves a 3.1 and 4.9\% performance improvement on the Youcook2 dataset using the CIDEr and F1 Score metrics; a 6.5\% and 3.7\% performance increase in Recall with IOU = \{0.5, 0.7\} thresholds on the Charades-STA dataset; and a 10.3\% and 9.2\% performance gain for the mAP and HIT@1 metrics on the QVHighlights dataset.
    \item \textbf{Better performance than task-specific models and larger LLMs.} As shown in Table~\ref{tab:zero-shot-vtg}, as a generalist model capable of handling various tasks, the performance of \alg surpasses that of task-specific models like HawkEye~\citep{wang2024hawkeye}. Furthermore, the 7B \alg model outperforms the VTimeLLM (13B) model~\citep{huang2023vtimellm}, further validating the advantages of the \alg architecture.
\end{itemize}
\begin{table*}[!t]
    \centering
    \caption{
    \small
        \textbf{Zero-shot performance of algorithms over various tasks.}
        We evaluated the performance of \alg using the Youcook2, Charades-STA, and QVHighlights datasets. 
        We highlight the best results for each block using \textbf{bold font}. The Valley, VideoChat-Embed, and Video-LLaMA results are elaborated from previous studies~\citep{ren2023timechat,huang2023vtimellm,qian2024momentor}. 
        The results with transparent text indicates unfair comparison (13B).
    }
    \vspace{-.5em}
    \setlength{\tabcolsep}{0.9mm}
    \fontsize{9pt}{11pt}\selectfont
    \resizebox{.9\textwidth}{!}{
        \begin{tabular}{l c c c c c c c c c}
            \toprule
            \multirow{2}{*}{Model} & \multicolumn{3}{c}{Youcook2}                         & \multicolumn{2}{c}{Charades-STA}                      & \multicolumn{2}{c}{QVHighlights}                                                                                                                                                                                                                                    \\
            \cmidrule(lr){2-4} \cmidrule(lr){5-6} \cmidrule(lr){7-8}
                                       & SODA\_c                                                          & CIDEr                                                         & F1 Score                                                          & $\text{R@1}_{\text{(IOU=0.5)}}$                                                         & $\text{R@1}_{\text{(IOU=0.7)}}$                                                                   & mAP & HIT@1                                                                 \\
            \midrule
            \textit{\textbf{\small Traditional Video LLMs}} \\
            Valley (7B) & 0.1 & 0.0 & 1.5 & 4.7 & 1.6 & 10.9 & 15.2                   \\
            VideoChat (7B) & 0.2 & 0.6 & 3.4 & 3.2 & 1.4 & 13.1 & 18.1 \\
            Video-LLaMA (7B) & 0.0 & 0.0 & 0.1 & 2.7 & 1.2 & 11.3 & 15.6 \\
            \midrule
            \textit{\textbf{\small Temporal Grounding Video LLMs}} \\
            TimeChat (7B) & 1.2 & 3.4 & 12.6 & 32.2 & 13.4 & 14.5 & 23.9\\
            VTimeLLM (7B) &  &  &  & 27.5 & 11.4 &  & \\
            {\texttransparent{0.5}{VTimeLLM (13B)}} &  &  &  & \texttransparent{0.5}{34.3} & \texttransparent{0.5}{14.7} & & \\
            Momentor (7B) &  &  &  & 26.6 & 11.6 & 7.6 &  \\
            HawkEye (7B) &  &  &  & 31.4 & 14.5 &  & \\
            VTG-LLM (7B) & 1.5 & 5.0 & 17.5 & 33.8 & 15.7 & 16.5 & 33.5 \\
            \midrule
            \alg (7B) & \textbf{2.2} & \textbf{8.1} & \textbf{22.4} & \textbf{40.3} & \textbf{19.4} & \textbf{26.8} & \textbf{42.7}\\
            \bottomrule
        \end{tabular}%
    }
    \vspace{-1em}
    \label{tab:zero-shot-vtg}
\end{table*}

\paragraph{Performance of \alg on ActivityNet Captions dataset.} In Table~\ref{tab:activitynet-performance}, we show the performance of \alg on ActivityNet Captions dataset. All the reported algorithms except for Momentor~\citep{qian2024momentor} have incorporated the ActivityNet Captions dataset as part of the training data. Results show that the \alg attains the best performance in moment retrieval tasks and demonstrates comparable results to VTimeLLM in dense video caption tasks.

\subsection{Ablation Studies of \alg.}

\paragraph{The \framework framework enhances model performance in VTG tasks.} In the 'Ablation Studies on Architecture' section of Table~\ref{tab:ablation}, we conducted experiments without utilizing the \framework framework. The results indicate that employing the \framework framework significantly improves model performance, and \alg can achieve better results even when sampling fewer video frames.

\paragraph{Using different encoders and decoding heads for different tasks is essential for \alg to achieve the best result.} In the "w/o independent ecoder/heads" part of Table~\ref{tab:ablation}, we performed ablation studies by not utilizing separate encoders and decoder heads for different tasks. Instead, we directly incorporated time tokens and score tokens into the text tokenizers. The results suggest that using shared encoder/decoding heads for \framework framework significantly disrupts the prelearned knowledge of LLMs, leading to irrelevant and meaningless responses.

\paragraph{The performance of \alg improves with the increase in the number of frames.} 
We conducted ablation studies on the number of sampled frames, as presented in Table~\ref{tab:ablation}. The results show that (1) the performance of \alg enhances as the number of sampled frames increases; (2) the performance of \alg is comparable or even superior to SOTA video LLMs like VTG-LLM and TimeChat when sampling just 8 frames, demonstrating the effectiveness of the \alg model architecture.

\begin{figure}[!t]
    \centering
    \subfigure[Performance on Youcook2]{
    \includegraphics[width=0.4\linewidth]{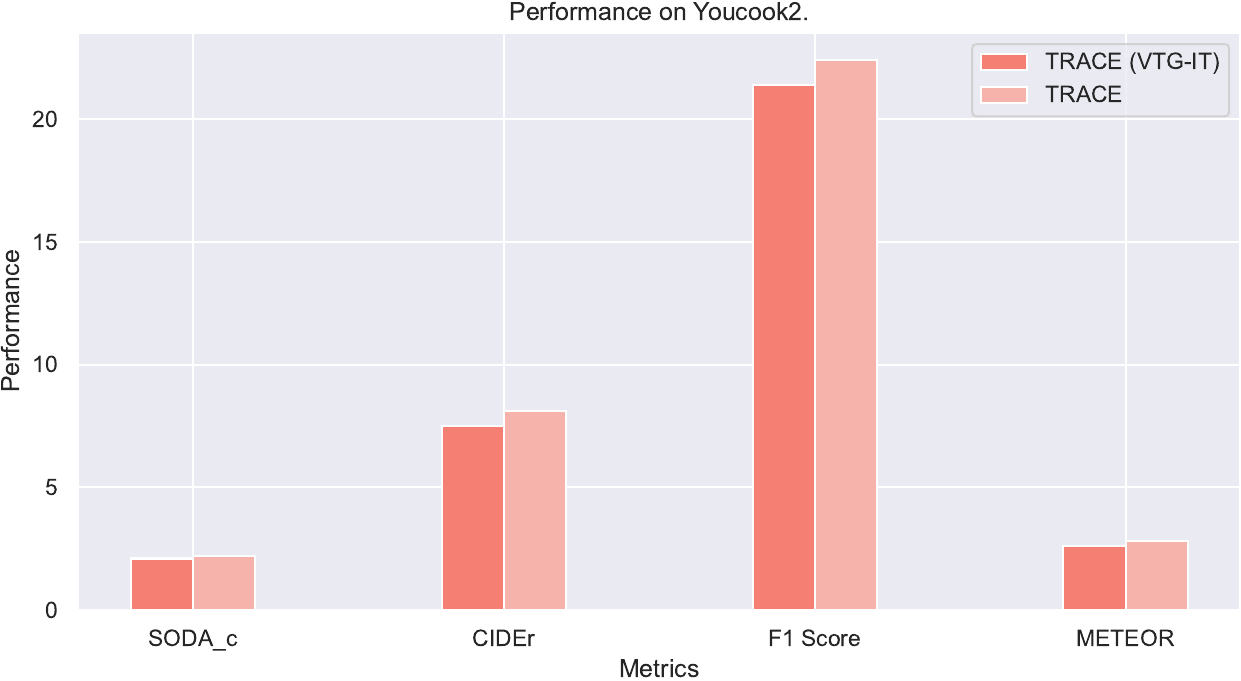}
    }
    \subfigure[Performance on Charades-STA]{
    \includegraphics[width=0.4\linewidth]{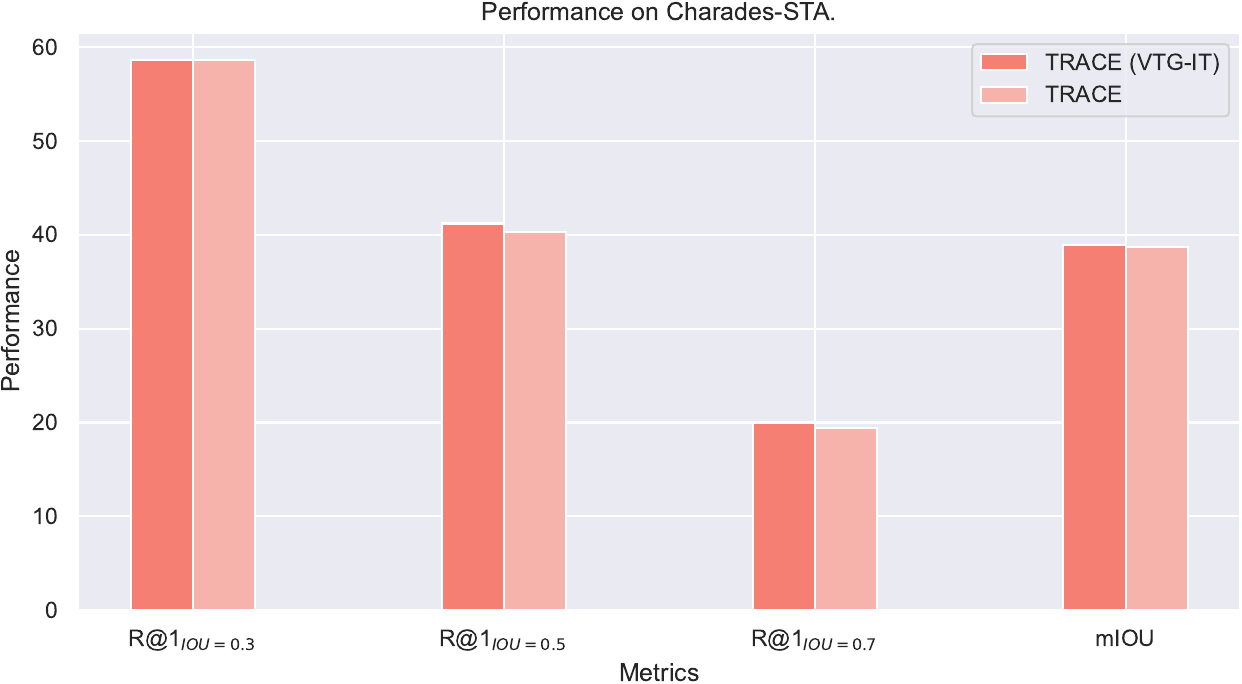}
    }
    \\
    \subfigure[Performance on QVHighlights]{
    \includegraphics[width=0.4\linewidth]{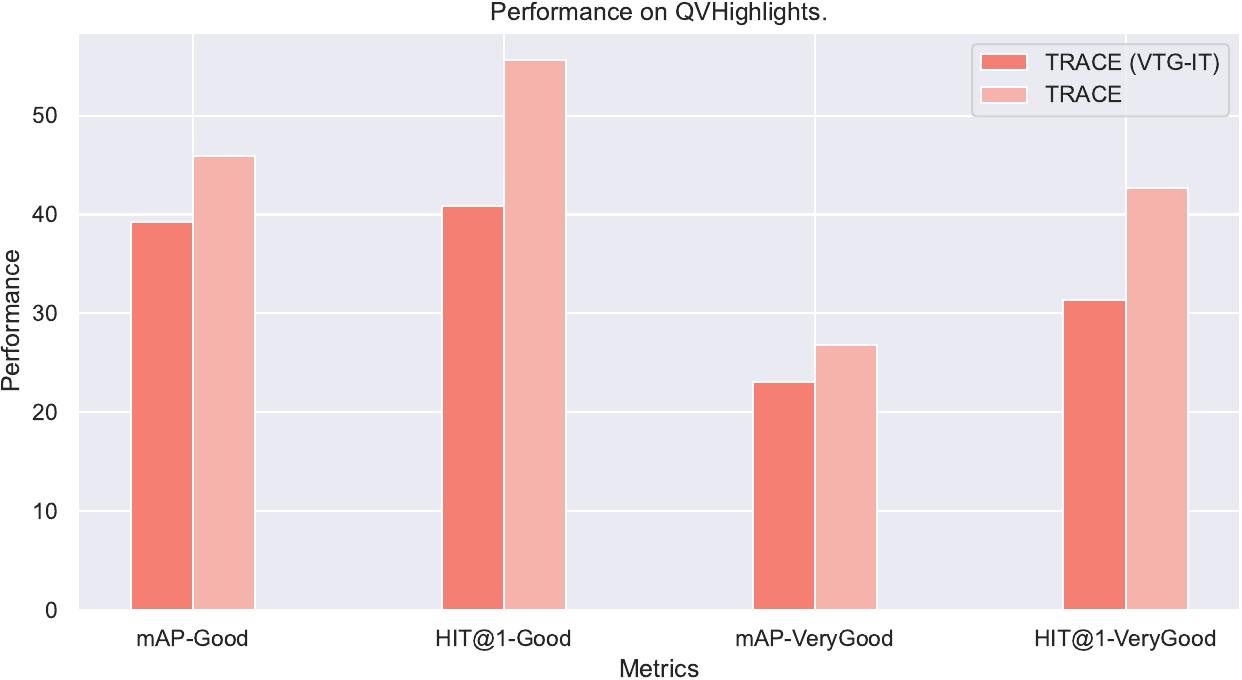}
    }
    \subfigure[Performance on ActivityNet Captions]{
    \includegraphics[width=0.4\linewidth]{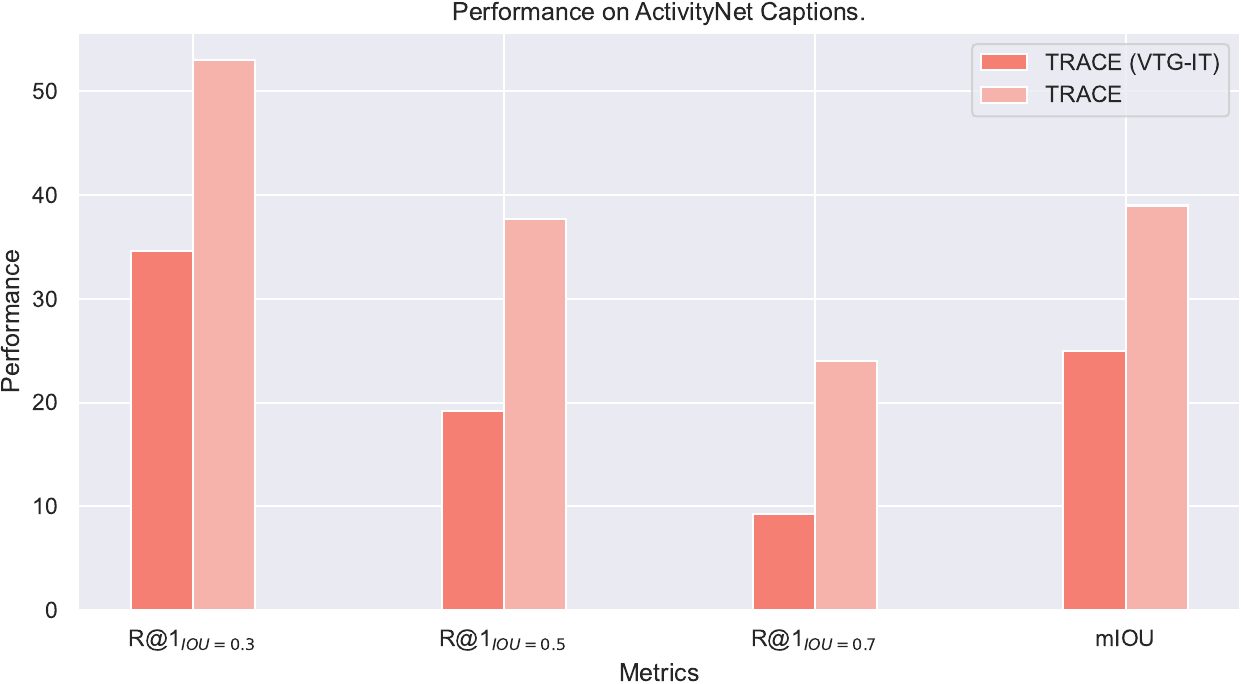}
    }
    \vspace{-1em}
    \caption{\small \textbf{Ablation studies on data utilized while training \alg.} We conduct experiments solely utilizing VTG-IT  and compare its performance with that of the original \alg.}
    \label{fig:ablation-data}
\end{figure}

\paragraph{Incorporating InternVid~\citep{wang2023internvid} and ActivityNet Captions~\citep{caba2015activitynet} datasets boost \alg performance on long videos.}  As illustrated in Figure~\ref{fig:ablation-data}, we carried out ablation studies by exclusively using VTG-IT as the training data for VTG tasks. The results indicate that the performance of \alg on long videos improves when incorporating internVid and ActivityNet Captions datasets, leading to enhanced performance on Youcook2, QVHighlights, and ActivityNet Captions datasets. Conversely, the performance of \alg on short videos slightly decreases (Charades-STA), suggesting that the annotations in the internVid and ActivityNet Captions datasets may not be as accurate as those in short video annotations.

\subsection{Fine-tuned Performance of \alg.} 

\paragraph{Competitive performance of \alg to traditional methods after fine-tuning.}
In Table~\ref{tab:fine-tune-performance}, we fine-tune the \alg for 3 epochs on Youcook2 and Charades-STA datasets~\footnote{Results on QVHighlights can be found in Appendix~\ref{sec:additional-experiments}.}. The results indicate that
\begin{itemize}[leftmargin=12pt,nosep]
    \item \textit{\alg significantly outperform generalist baselines.} In contrast to TimeChat and VTG-LLM, which struggle to attain satisfactory performance even after fine-tuning, the \alg derives significant benefits from fine-tuning and achieves notably better performance than generalist baselines. These results further substantiate that our enhancements to the model architecture are crucial for VTG tasks.
    \item \textit{\alg achieve comparable performance to non-generative and task-specific SOTAs.} As depicted in Table~\ref{tab:fine-tune-performance}, the \alg achieves new SOTA results on Youcook2 (without audio inputs). Furthermore, the performance of \alg on the Charades-STA dataset is also competitive with non-generative models such as InternVideo2 and VDI. \emph{However, these methods cannot handle various tasks simultaneously and lack zero-shot capability -- the contribution of \alg herein.}
\end{itemize}

\begin{table*}[!t]
    \centering
    \caption{
    \small
        \textbf{Ablation studies of \alg.} All the algorithms solely utilize VTG-IT~\citep{guo2024vtg} during fine-tuning for efficient evaluation. The "w/o \framework" approach indicates the use of natural language-style inputs similar to previous studies~\citep{guo2024vtg,ren2023timechat}. The "w/o independent encoder/heads" approach signifies directly adding new tokens to the LLM tokenizer instead of employing separate encoders/heads for different tasks. We highlight the best results using \textbf{bold font} for each block.
    }
    \vspace{-.5em}
    \setlength{\tabcolsep}{0.9mm}
    \fontsize{9pt}{11pt}\selectfont
    \resizebox{.9\textwidth}{!}{
        \begin{tabular}{l c c c c c c c c c c}
            \toprule
            \multirow{2}{*}{Models} & \multirow{2}{*}{Frame Number} &  \multicolumn{3}{c}{Youcook2}                         & \multicolumn{2}{c}{Charades-STA} \\
            \cmidrule(lr){3-5} \cmidrule(lr){6-7} 
                                       & & SODA\_c & CIDEr & F1 Score & $\text{R@1}_{\text{(IOU=0.5)}}$ & $\text{R@1}_{\text{(IOU=0.7)}}$  \\
            \midrule
            \textit{\textbf{Ablation Studies on Architecture}} \\
            w/o \framework & 96 & 1.4 & 4.3 & 17.2 & 29.7 & 14.0 \\
            w/o independent encoder/heads & 64 & \multicolumn{5}{c}{----Fail to Follow Instruction----}\\
            \alg (VTG-IT) & 64 & \textbf{1.9} & \textbf{6.9} & \textbf{21.4} & \textbf{37.0} & \textbf{17.0} \\
            \midrule
            \textit{\textbf{Ablation Studies on Frame Number}} \\
            \alg (VTG-IT) & 8 & 1.4 & 5.0 & 18.6 & 28.8 & 13.6 \\
            \alg (VTG-IT) & 64 & 1.9 & 6.9 & \textbf{21.4} & 37.0 & 17.0 \\
            \alg (VTG-IT) & 128 & \textbf{2.1} & \textbf{7.5} & \textbf{21.4} & \textbf{41.2} & \textbf{20.0} \\
            \bottomrule
        \end{tabular}%
    }
    \vspace{-.5em}
    \label{tab:ablation}
\end{table*}

\begin{table*}[!t]
    \centering
    \caption{
    \small
        \textbf{Performance of \alg on ActivityNet Captions dataset.} The evaluation of TimeChat's  and VTG-LLM's results was conducted using the official provided checkpoints. The $^{*}$ indicates zero-shot evaluation. We highlight the best and the second best results using \textbf{bold} and \underline{underline}.
    }
    \vspace{-.5em}
    \setlength{\tabcolsep}{0.9mm}
    \fontsize{9pt}{11pt}\selectfont
    \resizebox{.8\textwidth}{!}{
        \begin{tabular}{l c c c c c c c c c c}
            \toprule
            \multirow{2}{*}{Models} &  \multicolumn{4}{c}{Dense Video Caption}                         & \multicolumn{3}{c}{Moment Retrieval} \\
            \cmidrule(lr){2-5} \cmidrule(lr){6-8} 
                                       & METEOR & SODA\_c & CIDEr & F1 Score & $\text{R@1}_{\text{(IOU=0.5)}}$ & $\text{R@1}_{\text{(IOU=0.7)}}$ & mIOU \\
            \midrule
            VTimeLLM & \textbf{6.8} & \underline{5.8} & \textbf{27.6} & & \underline{29.5} & \underline{14.2} & \underline{31.4}\\
            Momentor$^{*}$ & 4.7 & 2.3 & 14.9 & & 23.0 & 12.4 & 29.3\\
            TimeChat & 5.7 & 4.7 & 19.0 & 36.9 & 4.6 & 2.0 & 6.9\\
            VTG-LLM & 5.9 & 5.1 & 20.7 & 34.8 & 8.3 & 3.7 & 12.0\\
            \midrule
            \alg & \underline{6.4} & \textbf{6.0} & \underline{25.9} & \textbf{39.3} & \textbf{37.7} & \textbf{24.0} & \textbf{39.0} \\
            \bottomrule
        \end{tabular}%
    }
    \label{tab:activitynet-performance}
\end{table*}

\begin{table}

 \caption{\small \textbf{Fine-tuned performance of \alg.} We fine-tune the \alg for 3 epochs on the Youcook2 and Charades-STA datasets. We emphasize the best and second best results using \textbf{bold font} and \underline{underline}. 
 For Youcook2 dataset, we choose Vid2Seq~\citep{yang2023vid2seq}, PDVC~\citep{wang2021end}, and CM$^2$~\citep{kim2024you} as task-specific baselines.
 The results depicted in \textcolor{lightgray}{gray} indicate unfair comparisons due to additional audio inputs and different architectures. 
 For charades-STA dataset, we choose InternVideo2-6B~\citep{wang2024internvideo2}, VDI~\citep{luo2023towards}, and Moment-DETR~\citep{lei2021detecting} as examples of non-generative models.}
 \label{tab:fine-tune-performance}

\begin{minipage}{.5\textwidth}
  \centering
  \setlength{\tabcolsep}{0.9mm}
    \fontsize{9pt}{11pt}\selectfont
  \resizebox{1.\textwidth}{!}{
  \begin{tabular}{l c c c c c c c c c}
            \toprule
            \multirow{2}{*}{Model} & \multicolumn{3}{c}{Youcook2}                         \\
            \cmidrule(lr){2-4}
                                       & SODA\_c                                                          & CIDEr                                                         & F1 Score                                                          \\
            \midrule
            \textit{Task-Specific Models} \\
            PDVC & 4.4 & 22.7 \\
            \textcolor{lightgray}{Vid2Seq (Audio Input)} & \textcolor{lightgray}{7.9} & \textcolor{lightgray}{ 47.1} & \textcolor{lightgray}{ 27.3}\\
            Vid2Seq & \underline{5.7} & 25.3 & 23.5 \\
            CM$^2$ & 5.3 & \underline{31.7} & \underline{28.4} \\
            
            \midrule
            \textit{Generalist Models} \\
            TimeChat & 3.4 & 11.0 & 19.5 \\
            VTG-LLM & 3.6 & 13.4 & 20.6 \\
            \alg & \textbf{6.7} & \textbf{35.5} & \textbf{31.8}\\
            \bottomrule
        \end{tabular}
        }
\end{minipage}%
\begin{minipage}{.5\textwidth}
  \centering
  \setlength{\tabcolsep}{0.9mm}
    \fontsize{9pt}{11pt}\selectfont
  \resizebox{1.\textwidth}{!}{
  \begin{tabular}{l c c c c c c c c c}
            \toprule
            \multirow{2}{*}{Model} & \multicolumn{2}{c}{Charades-STA}                         \\
            \cmidrule(lr){2-3}
                                       & $\text{R@1}_{\text{(IOU=0.5)}}$                                                         & $\text{R@1}_{\text{(IOU=0.7)}}$                                                         \\
            \midrule
            \textit{Non-Generative Models} \\
            \textcolor{lightgray}{InternVideo2-6B} &  \textcolor{lightgray}{70.0} & \textcolor{lightgray}{49.0} \\
            \textcolor{lightgray}{VDI}  & \textcolor{lightgray}{52.3} & \textcolor{lightgray}{31.4} \\
            \textcolor{lightgray}{Moment-DETR} & \textcolor{lightgray}{55.7} & \textcolor{lightgray}{34.2} \\
            \midrule
            \textit{Generative Models} \\
            HawkEye & \underline{58.3} & 28.8 \\
            TimeChat & 46.7 & 23.7 \\
            VTG-LLM  & 57.2 & \underline{33.4} \\
            \alg & \textbf{61.7} & \textbf{41.4} \\
            \bottomrule
        \end{tabular}
        }
\end{minipage}
\vspace{-1.5em}
\end{table}

\section{Conclusion and Future Works}

In this paper, our goal is to address the mismatch between video structure and video LLMs on VTG tasks,  and propose a \framework framework and the \alg model as a solution. Numerical results indicate the superior zero-shot performance of \alg compared to other video LLM baselines, and \alg also achieves competitive performance relative to traditional non-generative and task-specific models after fine-tuning. 
By overcoming the inherent limitations of video LLM architectures, \alg demonstrates the potential of video LLMs on VTG tasks, and we believe that the \alg could be a strong foundation for future research on video LLMs in VTG tasks.

However, there are future works that can further enhance the capabilities of \alg. 
{
For instance, \alg relies on the pre-trained decoder-only LLMs, and only using previous events to predict the next event, which may not discover the complex event relationships as pointed out by previous studies~\citep{yi2019clevrer,girdhar2019cater,li2020causal}, As a remedy, we can use the outputs of causality discovery models~\citep{liang2022visual,chen2024mecd} as supplementary inputs for TRACE to provide a more comprehensive understanding of video contents. 
Furthermore, expanding the annotation of more video understanding tasks by incorporating the occurrence timestamps of QA pairs and the matching score between questions and answers could significantly improve the overall performance of \alg.
}

\section*{Acknowledgments}
This work is supported in part by the funding from Shenzhen Institute of Artificial Intelligence and Robotics for Society, in part by the Shenzhen Key Lab of Crowd Intelligence Empowered Low-Carbon Energy Network (Grant No. ZDSYS20220606100601002), in part by Shenzhen Stability Science Program 2023, and in part by the Guangdong Provincial Key Laboratory of Future Networks of Intelligence (Grant No. 2022B1212010001).

\bibliography{iclr2025_conference}

\begin{thebibliography}{79}
\providecommand{\natexlab}[1]{#1}
\providecommand{\url}[1]{\texttt{#1}}
\expandafter\ifx\csname urlstyle\endcsname\relax
  \providecommand{\doi}[1]{doi: #1}\else
  \providecommand{\doi}{doi: \begingroup \urlstyle{rm}\Url}\fi

\bibitem[Achiam et~al.(2023)Achiam, Adler, Agarwal, Ahmad, Akkaya, Aleman,
  Almeida, Altenschmidt, Altman, Anadkat, et~al.]{achiam2023gpt}
Josh Achiam, Steven Adler, Sandhini Agarwal, Lama Ahmad, Ilge Akkaya,
  Florencia~Leoni Aleman, Diogo Almeida, Janko Altenschmidt, Sam Altman,
  Shyamal Anadkat, et~al.
\newblock Gpt-4 technical report.
\newblock \emph{arXiv preprint arXiv:2303.08774}, 2023.

\bibitem[Banerjee \& Lavie(2005)Banerjee and Lavie]{banerjee2005meteor}
Satanjeev Banerjee and Alon Lavie.
\newblock Meteor: An automatic metric for mt evaluation with improved
  correlation with human judgments.
\newblock In \emph{Proceedings of the acl workshop on intrinsic and extrinsic
  evaluation measures for machine translation and/or summarization}, pp.\
  65--72, 2005.

\bibitem[Boris et~al.(2024)Boris, Anil, Anna, and Marcus]{boris2024surprising}
Meinardus Boris, Batra Anil, Rohrbach Anna, and Rohrbach Marcus.
\newblock The surprising effectiveness of multimodal large language models for
  video moment retrieval.
\newblock \emph{arXiv preprint arXiv:2406.18113}, 2024.

\bibitem[Chen et~al.(2024{\natexlab{a}})Chen, Wei, Li, Dong, Zhang, Zang, Chen,
  Duan, Lin, Tang, et~al.]{chen2024sharegpt4video}
Lin Chen, Xilin Wei, Jinsong Li, Xiaoyi Dong, Pan Zhang, Yuhang Zang, Zehui
  Chen, Haodong Duan, Bin Lin, Zhenyu Tang, et~al.
\newblock Sharegpt4video: Improving video understanding and generation with
  better captions.
\newblock \emph{arXiv preprint arXiv:2406.04325}, 2024{\natexlab{a}}.

\bibitem[Chen et~al.(2024{\natexlab{b}})Chen, Li, Wang, Zhao, Sun, Zhu, and
  Liu]{chen2024vast}
Sihan Chen, Handong Li, Qunbo Wang, Zijia Zhao, Mingzhen Sun, Xinxin Zhu, and
  Jing Liu.
\newblock Vast: A vision-audio-subtitle-text omni-modality foundation model and
  dataset.
\newblock \emph{Advances in Neural Information Processing Systems}, 36,
  2024{\natexlab{b}}.

\bibitem[Chen et~al.(2024{\natexlab{c}})Chen, Liu, He, Chen, Gan, Ma, Zhong,
  Zhang, Wang, Lin, et~al.]{chen2024mecd}
Tieyuan Chen, Huabin Liu, Tianyao He, Yihang Chen, Chaofan Gan, Xiao Ma, Cheng
  Zhong, Yang Zhang, Yingxue Wang, Hui Lin, et~al.
\newblock Mecd: Unlocking multi-event causal discovery in video reasoning.
\newblock \emph{arXiv preprint arXiv:2409.17647}, 2024{\natexlab{c}}.

\bibitem[Cheng et~al.(2024)Cheng, Leng, Zhang, Xin, Li, Chen, Zhu, Zhang, Luo,
  Zhao, et~al.]{cheng2024videollama}
Zesen Cheng, Sicong Leng, Hang Zhang, Yifei Xin, Xin Li, Guanzheng Chen,
  Yongxin Zhu, Wenqi Zhang, Ziyang Luo, Deli Zhao, et~al.
\newblock Videollama 2: Advancing spatial-temporal modeling and audio
  understanding in video-llms.
\newblock \emph{arXiv preprint arXiv:2406.07476}, 2024.

\bibitem[Du et~al.(2024)Du, Zhou, Huo, Li, Zhao, Lu, Zhao, Wang, Chen, and
  Wen]{du2024towards}
Yifan Du, Kun Zhou, Yuqi Huo, Yifan Li, Wayne~Xin Zhao, Haoyu Lu, Zijia Zhao,
  Bingning Wang, Weipeng Chen, and Ji-Rong Wen.
\newblock Towards event-oriented long video understanding.
\newblock \emph{arXiv preprint arXiv:2406.14129}, 2024.

\bibitem[Fabian Caba~Heilbron \& Niebles(2015)Fabian Caba~Heilbron and
  Niebles]{caba2015activitynet}
Bernard~Ghanem Fabian Caba~Heilbron, Victor~Escorcia and Juan~Carlos Niebles.
\newblock Activitynet: A large-scale video benchmark for human activity
  understanding.
\newblock In \emph{Proceedings of the IEEE Conference on Computer Vision and
  Pattern Recognition}, pp.\  961--970, 2015.

\bibitem[Fujita et~al.(2020)Fujita, Hirao, Kamigaito, Okumura, and
  Nagata]{fujita2020soda}
Soichiro Fujita, Tsutomu Hirao, Hidetaka Kamigaito, Manabu Okumura, and Masaaki
  Nagata.
\newblock Soda: Story oriented dense video captioning evaluation framework.
\newblock In \emph{Computer Vision--ECCV 2020: 16th European Conference,
  Glasgow, UK, August 23--28, 2020, Proceedings, Part VI 16}, pp.\  517--531.
  Springer, 2020.

\bibitem[Gao et~al.(2017)Gao, Sun, Yang, and Nevatia]{gao2017tall}
Jiyang Gao, Chen Sun, Zhenheng Yang, and Ram Nevatia.
\newblock Tall: Temporal activity localization via language query.
\newblock In \emph{Proceedings of the IEEE international conference on computer
  vision}, pp.\  5267--5275, 2017.

\bibitem[Ghosal et~al.(2023)Ghosal, Majumder, Mehrish, and
  Poria]{ghosal2023text}
Deepanway Ghosal, Navonil Majumder, Ambuj Mehrish, and Soujanya Poria.
\newblock Text-to-audio generation using instruction-tuned llm and latent
  diffusion model.
\newblock \emph{arXiv preprint arXiv:2304.13731}, 2023.

\bibitem[Girdhar \& Ramanan(2019)Girdhar and Ramanan]{girdhar2019cater}
Rohit Girdhar and Deva Ramanan.
\newblock Cater: A diagnostic dataset for compositional actions and temporal
  reasoning.
\newblock \emph{arXiv preprint arXiv:1910.04744}, 2019.

\bibitem[Guo et~al.(2024)Guo, Liu, Li, Tang, Chen, and Zhao]{guo2024vtg}
Yongxin Guo, Jingyu Liu, Mingda Li, Xiaoying Tang, Xi~Chen, and Bo~Zhao.
\newblock Vtg-llm: Integrating timestamp knowledge into video llms for enhanced
  video temporal grounding.
\newblock \emph{arXiv preprint arXiv:2405.13382}, 2024.

\bibitem[Gygli et~al.(2014)Gygli, Grabner, Riemenschneider, and
  Van~Gool]{gygli2014creating}
Michael Gygli, Helmut Grabner, Hayko Riemenschneider, and Luc Van~Gool.
\newblock Creating summaries from user videos.
\newblock In \emph{Computer Vision--ECCV 2014: 13th European Conference,
  Zurich, Switzerland, September 6-12, 2014, Proceedings, Part VII 13}, pp.\
  505--520. Springer, 2014.

\bibitem[Han et~al.(2024)Han, Seo, Park, Nam, and Kwak]{han2024unleash}
Donghoon Han, Seunghyeon Seo, Eunhwan Park, Seong-Uk Nam, and Nojun Kwak.
\newblock Unleash the potential of clip for video highlight detection.
\newblock \emph{arXiv preprint arXiv:2404.01745}, 2024.

\bibitem[Hendricks et~al.(2018{\natexlab{a}})Hendricks, Wang, Shechtman, Sivic,
  Darrell, and Russell]{hendricks18emnlp}
Lisa~Anne Hendricks, Oliver Wang, Eli Shechtman, Josef Sivic, Trevor Darrell,
  and Bryan Russell.
\newblock Localizing moments in video with temporal language.
\newblock In \emph{Empirical Methods in Natural Language Processing (EMNLP)},
  2018{\natexlab{a}}.

\bibitem[Hendricks et~al.(2018{\natexlab{b}})Hendricks, Wang, Shechtman, Sivic,
  Darrell, and Russell]{hendricks2018localizing}
Lisa~Anne Hendricks, Oliver Wang, Eli Shechtman, Josef Sivic, Trevor Darrell,
  and Bryan Russell.
\newblock Localizing moments in video with temporal language.
\newblock In \emph{Empirical Methods in Natural Language Processing (EMNLP)},
  2018{\natexlab{b}}.

\bibitem[Hua et~al.(2024)Hua, Tang, Xu, and Luo]{hua2024v2xum}
Hang Hua, Yunlong Tang, Chenliang Xu, and Jiebo Luo.
\newblock V2xum-llm: Cross-modal video summarization with temporal prompt
  instruction tuning.
\newblock \emph{arXiv preprint arXiv:2404.12353}, 2024.

\bibitem[Huang et~al.(2023)Huang, Wang, Chen, Song, and Zhu]{huang2023vtimellm}
Bin Huang, Xin Wang, Hong Chen, Zihan Song, and Wenwu Zhu.
\newblock Vtimellm: Empower llm to grasp video moments.
\newblock \emph{arXiv preprint arXiv:2311.18445}, 2\penalty0 (3):\penalty0 9,
  2023.

\bibitem[Huang et~al.(2024)Huang, Liao, Radhakrishnan, Yin, Molchanov, Yu, and
  Kautz]{huang2024lita}
De-An Huang, Shijia Liao, Subhashree Radhakrishnan, Hongxu Yin, Pavlo
  Molchanov, Zhiding Yu, and Jan Kautz.
\newblock Lita: Language instructed temporal-localization assistant.
\newblock \emph{arXiv preprint arXiv:2403.19046}, 2024.

\bibitem[Jiang et~al.(2023)Jiang, Sablayrolles, Mensch, Bamford, Chaplot,
  Casas, Bressand, Lengyel, Lample, Saulnier, et~al.]{jiang2023mistral}
Albert~Q Jiang, Alexandre Sablayrolles, Arthur Mensch, Chris Bamford,
  Devendra~Singh Chaplot, Diego de~las Casas, Florian Bressand, Gianna Lengyel,
  Guillaume Lample, Lucile Saulnier, et~al.
\newblock Mistral 7b.
\newblock \emph{arXiv preprint arXiv:2310.06825}, 2023.

\bibitem[Jin et~al.(2022)Jin, Zhu, and Mu]{jin2022complex}
Yang Jin, Linchao Zhu, and Yadong Mu.
\newblock Complex video action reasoning via learnable markov logic network.
\newblock In \emph{Proceedings of the IEEE/CVF Conference on Computer Vision
  and Pattern Recognition}, pp.\  3242--3251, 2022.

\bibitem[Kaplan et~al.(2020)Kaplan, McCandlish, Henighan, Brown, Chess, Child,
  Gray, Radford, Wu, and Amodei]{kaplan2020scaling}
Jared Kaplan, Sam McCandlish, Tom Henighan, Tom~B Brown, Benjamin Chess, Rewon
  Child, Scott Gray, Alec Radford, Jeffrey Wu, and Dario Amodei.
\newblock Scaling laws for neural language models.
\newblock \emph{arXiv preprint arXiv:2001.08361}, 2020.

\bibitem[Kim et~al.(2024)Kim, Kim, Moon, Choi, and Kim]{kim2024you}
Minkuk Kim, Hyeon~Bae Kim, Jinyoung Moon, Jinwoo Choi, and Seong~Tae Kim.
\newblock Do you remember? dense video captioning with cross-modal memory
  retrieval.
\newblock \emph{arXiv preprint arXiv:2404.07610}, 2024.

\bibitem[Lei et~al.()Lei, Berg, and Bansal]{lei2107qvhighlights}
J~Lei, TL~Berg, and M~Bansal.
\newblock Qvhighlights: Detecting moments and highlights in videos via natural
  language queries.(2021).
\newblock \emph{URL https://arxiv. org/abs/2107.09609}.

\bibitem[Lei et~al.(2021)Lei, Berg, and Bansal]{lei2021detecting}
Jie Lei, Tamara~L Berg, and Mohit Bansal.
\newblock Detecting moments and highlights in videos via natural language
  queries.
\newblock \emph{Advances in Neural Information Processing Systems},
  34:\penalty0 11846--11858, 2021.

\bibitem[Li et~al.(2024)Li, Zhang, Guo, Zhang, Li, Zhang, Zhang, Li, Liu, and
  Li]{li2024llava}
Bo~Li, Yuanhan Zhang, Dong Guo, Renrui Zhang, Feng Li, Hao Zhang, Kaichen
  Zhang, Yanwei Li, Ziwei Liu, and Chunyuan Li.
\newblock Llava-onevision: Easy visual task transfer.
\newblock \emph{arXiv preprint arXiv:2408.03326}, 2024.

\bibitem[Li et~al.(2023{\natexlab{a}})Li, Li, Savarese, and Hoi]{li2023blip}
Junnan Li, Dongxu Li, Silvio Savarese, and Steven Hoi.
\newblock Blip-2: Bootstrapping language-image pre-training with frozen image
  encoders and large language models.
\newblock In \emph{International conference on machine learning}, pp.\
  19730--19742. PMLR, 2023{\natexlab{a}}.

\bibitem[Li et~al.(2023{\natexlab{b}})Li, He, Wang, Li, Wang, Luo, Wang, Wang,
  and Qiao]{li2023videochat}
Kunchang Li, Yinan He, Yi~Wang, Yizhuo Li, Wenhai Wang, Ping Luo, Yali Wang,
  Limin Wang, and Yu~Qiao.
\newblock Videochat: Chat-centric video understanding.
\newblock \emph{arXiv preprint arXiv:2305.06355}, 2023{\natexlab{b}}.

\bibitem[Li et~al.(2023{\natexlab{c}})Li, Wang, He, Li, Wang, Liu, Wang, Xu,
  Chen, Luo, Wang, and Qiao]{li2023mvbench}
Kunchang Li, Yali Wang, Yinan He, Yizhuo Li, Yi~Wang, Yi~Liu, Zun Wang, Jilan
  Xu, Guo Chen, Ping Luo, Limin Wang, and Yu~Qiao.
\newblock Mvbench: A comprehensive multi-modal video understanding benchmark,
  2023{\natexlab{c}}.

\bibitem[Li et~al.(2023{\natexlab{d}})Li, Wang, Li, Wang, He, Wang, and
  Qiao]{li2023unmasked}
Kunchang Li, Yali Wang, Yizhuo Li, Yi~Wang, Yinan He, Limin Wang, and Yu~Qiao.
\newblock Unmasked teacher: Towards training-efficient video foundation models.
\newblock In \emph{Proceedings of the IEEE/CVF International Conference on
  Computer Vision}, pp.\  19948--19960, 2023{\natexlab{d}}.

\bibitem[Li et~al.(2020)Li, Torralba, Anandkumar, Fox, and Garg]{li2020causal}
Yunzhu Li, Antonio Torralba, Anima Anandkumar, Dieter Fox, and Animesh Garg.
\newblock Causal discovery in physical systems from videos.
\newblock \emph{Advances in Neural Information Processing Systems},
  33:\penalty0 9180--9192, 2020.

\bibitem[Liang et~al.(2022)Liang, Wang, Zhou, and Yang]{liang2022visual}
Chen Liang, Wenguan Wang, Tianfei Zhou, and Yi~Yang.
\newblock Visual abductive reasoning.
\newblock In \emph{Proceedings of the IEEE/CVF conference on computer vision
  and pattern recognition}, pp.\  15565--15575, 2022.

\bibitem[Lin et~al.(2023{\natexlab{a}})Lin, Zhu, Ye, Ning, Jin, and
  Yuan]{lin2023video}
Bin Lin, Bin Zhu, Yang Ye, Munan Ning, Peng Jin, and Li~Yuan.
\newblock Video-llava: Learning united visual representation by alignment
  before projection.
\newblock \emph{arXiv preprint arXiv:2311.10122}, 2023{\natexlab{a}}.

\bibitem[Lin et~al.(2023{\natexlab{b}})Lin, Zhang, Chen, Pramanick, Gao, Wang,
  Yan, and Shou]{lin2023univtg}
Kevin~Qinghong Lin, Pengchuan Zhang, Joya Chen, Shraman Pramanick, Difei Gao,
  Alex~Jinpeng Wang, Rui Yan, and Mike~Zheng Shou.
\newblock Univtg: Towards unified video-language temporal grounding.
\newblock In \emph{Proceedings of the IEEE/CVF International Conference on
  Computer Vision}, pp.\  2794--2804, 2023{\natexlab{b}}.

\bibitem[Liu et~al.(2024)Liu, Li, Wu, and Lee]{liu2024visual}
Haotian Liu, Chunyuan Li, Qingyang Wu, and Yong~Jae Lee.
\newblock Visual instruction tuning.
\newblock \emph{Advances in neural information processing systems}, 36, 2024.

\bibitem[Liu et~al.()Liu, Ma, Qi, Wu, Shan, and Chen]{liu2024bench}
Ye~Liu, Zongyang Ma, Zhongang Qi, Yang Wu, Ying Shan, and Chang~Wen Chen.
\newblock Et bench: Towards open-ended event-level video-language
  understanding.
\newblock In \emph{The Thirty-eight Conference on Neural Information Processing
  Systems Datasets and Benchmarks Track}.

\bibitem[Liu et~al.(2022)Liu, Li, Wu, Chen, Shan, and Qie]{liu2022umt}
Ye~Liu, Siyuan Li, Yang Wu, Chang-Wen Chen, Ying Shan, and Xiaohu Qie.
\newblock Umt: Unified multi-modal transformers for joint video moment
  retrieval and highlight detection.
\newblock In \emph{Proceedings of the IEEE/CVF Conference on Computer Vision
  and Pattern Recognition}, pp.\  3042--3051, 2022.

\bibitem[Luo et~al.(2023{\natexlab{a}})Luo, Huang, Gong, Jin, and
  Liu]{luo2023towards}
Dezhao Luo, Jiabo Huang, Shaogang Gong, Hailin Jin, and Yang Liu.
\newblock Towards generalisable video moment retrieval: Visual-dynamic
  injection to image-text pre-training.
\newblock In \emph{Proceedings of the IEEE/CVF Conference on Computer Vision
  and Pattern Recognition}, pp.\  23045--23055, 2023{\natexlab{a}}.

\bibitem[Luo et~al.(2023{\natexlab{b}})Luo, Zhao, Yang, Dong, Qiu, Lu, Wang,
  and Wei]{luo2023valley}
Ruipu Luo, Ziwang Zhao, Min Yang, Junwei Dong, Minghui Qiu, Pengcheng Lu, Tao
  Wang, and Zhongyu Wei.
\newblock Valley: Video assistant with large language model enhanced ability.
\newblock \emph{arXiv preprint arXiv:2306.07207}, 2023{\natexlab{b}}.

\bibitem[Maaz et~al.(2023)Maaz, Rasheed, Khan, and Khan]{maaz2023video}
Muhammad Maaz, Hanoona Rasheed, Salman Khan, and Fahad~Shahbaz Khan.
\newblock Video-chatgpt: Towards detailed video understanding via large vision
  and language models.
\newblock \emph{arXiv preprint arXiv:2306.05424}, 2023.

\bibitem[Oncescu et~al.(2021)Oncescu, Henriques, Liu, Zisserman, and
  Albanie]{oncescu2021queryd}
Andreea-Maria Oncescu, Joao~F Henriques, Yang Liu, Andrew Zisserman, and Samuel
  Albanie.
\newblock Queryd: A video dataset with high-quality text and audio narrations.
\newblock In \emph{ICASSP 2021-2021 IEEE International Conference on Acoustics,
  Speech and Signal Processing (ICASSP)}, pp.\  2265--2269. IEEE, 2021.

\bibitem[Parmar et~al.(2024)Parmar, Peh, Chen, Lam, Chen, Tan, and
  Fernando]{parmar2024causalchaos}
Paritosh Parmar, Eric Peh, Ruirui Chen, Ting~En Lam, Yuhan Chen, Elston Tan,
  and Basura Fernando.
\newblock Causalchaos! dataset for comprehensive causal action question
  answering over longer causal chains grounded in dynamic visual scenes.
\newblock \emph{arXiv preprint arXiv:2404.01299}, 2024.

\bibitem[Qian et~al.(2024)Qian, Li, Wu, Ye, Fei, Chua, Zhuang, and
  Tang]{qian2024momentor}
Long Qian, Juncheng Li, Yu~Wu, Yaobo Ye, Hao Fei, Tat-Seng Chua, Yueting
  Zhuang, and Siliang Tang.
\newblock Momentor: Advancing video large language model with fine-grained
  temporal reasoning, 2024.

\bibitem[Radford et~al.(2021)Radford, Kim, Hallacy, Ramesh, Goh, Agarwal,
  Sastry, Askell, Mishkin, Clark, et~al.]{radford2021learning}
Alec Radford, Jong~Wook Kim, Chris Hallacy, Aditya Ramesh, Gabriel Goh,
  Sandhini Agarwal, Girish Sastry, Amanda Askell, Pamela Mishkin, Jack Clark,
  et~al.
\newblock Learning transferable visual models from natural language
  supervision.
\newblock In \emph{International conference on machine learning}, pp.\
  8748--8763. PMLR, 2021.

\bibitem[Ren et~al.(2023)Ren, Yao, Li, Sun, and Hou]{ren2023timechat}
Shuhuai Ren, Linli Yao, Shicheng Li, Xu~Sun, and Lu~Hou.
\newblock Timechat: A time-sensitive multimodal large language model for long
  video understanding.
\newblock \emph{arXiv preprint arXiv:2312.02051}, 2023.

\bibitem[Song et~al.(2024{\natexlab{a}})Song, Chai, Wang, Zhang, Zhou, Wu, Chi,
  Guo, Ye, Zhang, et~al.]{song2024moviechat}
Enxin Song, Wenhao Chai, Guanhong Wang, Yucheng Zhang, Haoyang Zhou, Feiyang
  Wu, Haozhe Chi, Xun Guo, Tian Ye, Yanting Zhang, et~al.
\newblock Moviechat: From dense token to sparse memory for long video
  understanding.
\newblock In \emph{Proceedings of the IEEE/CVF Conference on Computer Vision
  and Pattern Recognition}, pp.\  18221--18232, 2024{\natexlab{a}}.

\bibitem[Song et~al.(2024{\natexlab{b}})Song, Chai, Ye, Hwang, Li, and
  Wang]{song2024moviechat+}
Enxin Song, Wenhao Chai, Tian Ye, Jenq-Neng Hwang, Xi~Li, and Gaoang Wang.
\newblock Moviechat+: Question-aware sparse memory for long video question
  answering.
\newblock \emph{arXiv preprint arXiv:2404.17176}, 2024{\natexlab{b}}.

\bibitem[Song et~al.(2015)Song, Vallmitjana, Stent, and Jaimes]{song2015tvsum}
Yale Song, Jordi Vallmitjana, Amanda Stent, and Alejandro Jaimes.
\newblock Tvsum: Summarizing web videos using titles.
\newblock In \emph{Proceedings of the IEEE conference on computer vision and
  pattern recognition}, pp.\  5179--5187, 2015.

\bibitem[Sun et~al.(2023)Sun, Fang, Wu, Wang, and Cao]{sun2023eva}
Quan Sun, Yuxin Fang, Ledell Wu, Xinlong Wang, and Yue Cao.
\newblock Eva-clip: Improved training techniques for clip at scale.
\newblock \emph{arXiv preprint arXiv:2303.15389}, 2023.

\bibitem[Tang et~al.(2019)Tang, Ding, Rao, Zheng, Zhang, Zhao, Lu, and
  Zhou]{tang2019coin}
Yansong Tang, Dajun Ding, Yongming Rao, Yu~Zheng, Danyang Zhang, Lili Zhao,
  Jiwen Lu, and Jie Zhou.
\newblock Coin: A large-scale dataset for comprehensive instructional video
  analysis.
\newblock In \emph{Proceedings of the IEEE/CVF Conference on Computer Vision
  and Pattern Recognition}, pp.\  1207--1216, 2019.

\bibitem[Tong et~al.(2022)Tong, Song, Wang, and Wang]{tong2022videomae}
Zhan Tong, Yibing Song, Jue Wang, and Limin Wang.
\newblock Videomae: Masked autoencoders are data-efficient learners for
  self-supervised video pre-training.
\newblock \emph{Advances in neural information processing systems},
  35:\penalty0 10078--10093, 2022.

\bibitem[Touvron et~al.(2023)Touvron, Lavril, Izacard, Martinet, Lachaux,
  Lacroix, Rozi{\`e}re, Goyal, Hambro, Azhar, et~al.]{touvron2023llama}
Hugo Touvron, Thibaut Lavril, Gautier Izacard, Xavier Martinet, Marie-Anne
  Lachaux, Timoth{\'e}e Lacroix, Baptiste Rozi{\`e}re, Naman Goyal, Eric
  Hambro, Faisal Azhar, et~al.
\newblock Llama: Open and efficient foundation language models.
\newblock \emph{arXiv preprint arXiv:2302.13971}, 2023.

\bibitem[Vedantam et~al.(2015)Vedantam, Lawrence~Zitnick, and
  Parikh]{vedantam2015cider}
Ramakrishna Vedantam, C~Lawrence~Zitnick, and Devi Parikh.
\newblock Cider: Consensus-based image description evaluation.
\newblock In \emph{Proceedings of the IEEE conference on computer vision and
  pattern recognition}, pp.\  4566--4575, 2015.

\bibitem[Wang et~al.(2021)Wang, Zhang, Lu, Zheng, Cheng, and Luo]{wang2021end}
Teng Wang, Ruimao Zhang, Zhichao Lu, Feng Zheng, Ran Cheng, and Ping Luo.
\newblock End-to-end dense video captioning with parallel decoding.
\newblock In \emph{Proceedings of the IEEE/CVF international conference on
  computer vision}, pp.\  6847--6857, 2021.

\bibitem[Wang et~al.(2023{\natexlab{a}})Wang, Zhang, Zheng, Jiang, Cheng, and
  Luo]{wang2023learning}
Teng Wang, Jinrui Zhang, Feng Zheng, Wenhao Jiang, Ran Cheng, and Ping Luo.
\newblock Learning grounded vision-language representation for versatile
  understanding in untrimmed videos.
\newblock \emph{arXiv preprint arXiv:2303.06378}, 2023{\natexlab{a}}.

\bibitem[Wang et~al.(2022)Wang, Li, Li, He, Huang, Zhao, Zhang, Xu, Liu, Wang,
  et~al.]{wang2022internvideo}
Yi~Wang, Kunchang Li, Yizhuo Li, Yinan He, Bingkun Huang, Zhiyu Zhao, Hongjie
  Zhang, Jilan Xu, Yi~Liu, Zun Wang, et~al.
\newblock Internvideo: General video foundation models via generative and
  discriminative learning.
\newblock \emph{arXiv preprint arXiv:2212.03191}, 2022.

\bibitem[Wang et~al.(2023{\natexlab{b}})Wang, He, Li, Li, Yu, Ma, Li, Chen,
  Chen, Wang, et~al.]{wang2023internvid}
Yi~Wang, Yinan He, Yizhuo Li, Kunchang Li, Jiashuo Yu, Xin Ma, Xinhao Li, Guo
  Chen, Xinyuan Chen, Yaohui Wang, et~al.
\newblock Internvid: A large-scale video-text dataset for multimodal
  understanding and generation.
\newblock \emph{arXiv preprint arXiv:2307.06942}, 2023{\natexlab{b}}.

\bibitem[Wang et~al.(2024{\natexlab{a}})Wang, Li, Li, Yu, He, Chen, Pei, Zheng,
  Xu, Wang, et~al.]{wang2024internvideo2}
Yi~Wang, Kunchang Li, Xinhao Li, Jiashuo Yu, Yinan He, Guo Chen, Baoqi Pei,
  Rongkun Zheng, Jilan Xu, Zun Wang, et~al.
\newblock Internvideo2: Scaling video foundation models for multimodal video
  understanding.
\newblock \emph{arXiv preprint arXiv:2403.15377}, 2024{\natexlab{a}}.

\bibitem[Wang et~al.(2024{\natexlab{b}})Wang, Meng, Liang, Wang, Liu, and
  Zhao]{wang2024hawkeye}
Yueqian Wang, Xiaojun Meng, Jianxin Liang, Yuxuan Wang, Qun Liu, and Dongyan
  Zhao.
\newblock Hawkeye: Training video-text llms for grounding text in videos.
\newblock \emph{arXiv preprint arXiv:2403.10228}, 2024{\natexlab{b}}.

\bibitem[Wang et~al.(2024{\natexlab{c}})Wang, Wang, Wu, Liang, Zhao, Liu, and
  Zheng]{wang2024efficient}
Yuxuan Wang, Yueqian Wang, Pengfei Wu, Jianxin Liang, Dongyan Zhao, Yang Liu,
  and Zilong Zheng.
\newblock Efficient temporal extrapolation of multimodal large language models
  with temporal grounding bridge.
\newblock \emph{arXiv preprint arXiv:2402.16050}, 2024{\natexlab{c}}.

\bibitem[Wu et~al.(2025)Wu, Zhao, Li, Li, Zhou, Shou, and Bai]{wu2025large}
Weijia Wu, Yuzhong Zhao, Zhuang Li, Jiahong Li, Hong Zhou, Mike~Zheng Shou, and
  Xiang Bai.
\newblock A large cross-modal video retrieval dataset with reading
  comprehension.
\newblock \emph{Pattern Recognition}, 157:\penalty0 110818, 2025.

\bibitem[Wu et~al.(2024)Wu, Hu, Sun, Zhou, Zhu, Rao, Schiele, and
  Yang]{wu2024number}
Yongliang Wu, Xinting Hu, Yuyang Sun, Yizhou Zhou, Wenbo Zhu, Fengyun Rao,
  Bernt Schiele, and Xu~Yang.
\newblock Number it: Temporal grounding videos like flipping manga.
\newblock \emph{arXiv preprint arXiv:2411.10332}, 2024.

\bibitem[Xiao et~al.(2021)Xiao, Shang, Yao, and Chua]{xiao2021next}
Junbin Xiao, Xindi Shang, Angela Yao, and Tat-Seng Chua.
\newblock Next-qa: Next phase of question-answering to explaining temporal
  actions.
\newblock In \emph{Proceedings of the IEEE/CVF conference on computer vision
  and pattern recognition}, pp.\  9777--9786, 2021.

\bibitem[Xiao et~al.(2023)Xiao, Luo, Liu, Ma, Bian, Ji, Yang, and
  Li]{xiao2023bridging}
Yicheng Xiao, Zhuoyan Luo, Yong Liu, Yue Ma, Hengwei Bian, Yatai Ji, Yujiu
  Yang, and Xiu Li.
\newblock Bridging the gap: A unified video comprehension framework for moment
  retrieval and highlight detection.
\newblock \emph{arXiv preprint arXiv:2311.16464}, 2023.

\bibitem[Xu et~al.(2021)Xu, Ghosh, Huang, Okhonko, Aghajanyan, Metze,
  Zettlemoyer, and Feichtenhofer]{xu2021videoclip}
Hu~Xu, Gargi Ghosh, Po-Yao Huang, Dmytro Okhonko, Armen Aghajanyan, Florian
  Metze, Luke Zettlemoyer, and Christoph Feichtenhofer.
\newblock Videoclip: Contrastive pre-training for zero-shot video-text
  understanding.
\newblock \emph{arXiv preprint arXiv:2109.14084}, 2021.

\bibitem[Yan et~al.(2022)Yan, Zhu, Wang, Cao, Zhang, Ghosh, Wu, and
  Yu]{yan2022videococa}
Shen Yan, Tao Zhu, Zirui Wang, Yuan Cao, Mi~Zhang, Soham Ghosh, Yonghui Wu, and
  Jiahui Yu.
\newblock Videococa: Video-text modeling with zero-shot transfer from
  contrastive captioners.
\newblock \emph{arXiv preprint arXiv:2212.04979}, 2022.

\bibitem[Yang et~al.(2023)Yang, Nagrani, Seo, Miech, Pont-Tuset, Laptev, Sivic,
  and Schmid]{yang2023vid2seq}
Antoine Yang, Arsha Nagrani, Paul~Hongsuck Seo, Antoine Miech, Jordi
  Pont-Tuset, Ivan Laptev, Josef Sivic, and Cordelia Schmid.
\newblock Vid2seq: Large-scale pretraining of a visual language model for dense
  video captioning.
\newblock In \emph{Proceedings of the IEEE/CVF Conference on Computer Vision
  and Pattern Recognition}, pp.\  10714--10726, 2023.

\bibitem[Yao et~al.(2024)Yao, Yu, Zhang, Wang, Cui, Zhu, Cai, Li, Zhao, He,
  et~al.]{yao2024minicpm}
Yuan Yao, Tianyu Yu, Ao~Zhang, Chongyi Wang, Junbo Cui, Hongji Zhu, Tianchi
  Cai, Haoyu Li, Weilin Zhao, Zhihui He, et~al.
\newblock Minicpm-v: A gpt-4v level mllm on your phone.
\newblock \emph{arXiv preprint arXiv:2408.01800}, 2024.

\bibitem[Yi et~al.(2019)Yi, Gan, Li, Kohli, Wu, Torralba, and
  Tenenbaum]{yi2019clevrer}
Kexin Yi, Chuang Gan, Yunzhu Li, Pushmeet Kohli, Jiajun Wu, Antonio Torralba,
  and Joshua~B Tenenbaum.
\newblock Clevrer: Collision events for video representation and reasoning.
\newblock \emph{arXiv preprint arXiv:1910.01442}, 2019.

\bibitem[Zala et~al.(2023{\natexlab{a}})Zala, Cho, Kottur, Chen, Oguz, Mehdad,
  and Bansal]{zala2023hierarchical}
Abhay Zala, Jaemin Cho, Satwik Kottur, Xilun Chen, Barlas Oguz, Yashar Mehdad,
  and Mohit Bansal.
\newblock Hierarchical video-moment retrieval and step-captioning.
\newblock In \emph{Proceedings of the IEEE/CVF Conference on Computer Vision
  and Pattern Recognition}, pp.\  23056--23065, 2023{\natexlab{a}}.

\bibitem[Zala et~al.(2023{\natexlab{b}})Zala, Cho, Kottur, Chen, Oğuz, Mehdad,
  and Bansal]{Zala2023HiREST}
Abhay Zala, Jaemin Cho, Satwik Kottur, Xilun Chen, Barlas Oğuz, Yashar Mehdad,
  and Mohit Bansal.
\newblock Hierarchical video-moment retrieval and step-captioning.
\newblock In \emph{CVPR}, 2023{\natexlab{b}}.

\bibitem[Zellers et~al.(2021)Zellers, Lu, Hessel, Yu, Park, Cao, Farhadi, and
  Choi]{zellersluhessel2021merlot}
Rowan Zellers, Ximing Lu, Jack Hessel, Youngjae Yu, Jae~Sung Park, Jize Cao,
  Ali Farhadi, and Yejin Choi.
\newblock Merlot: Multimodal neural script knowledge models.
\newblock In \emph{Advances in Neural Information Processing Systems 34}, 2021.

\bibitem[Zhang et~al.(2023)Zhang, Li, and Bing]{zhang2023video}
Hang Zhang, Xin Li, and Lidong Bing.
\newblock Video-llama: An instruction-tuned audio-visual language model for
  video understanding.
\newblock \emph{arXiv preprint arXiv:2306.02858}, 2023.

\bibitem[Zhang et~al.(2024)Zhang, Wu, Li, Li, Ma, Liu, and Li]{zhang2024video}
Yuanhan Zhang, Jinming Wu, Wei Li, Bo~Li, Zejun Ma, Ziwei Liu, and Chunyuan Li.
\newblock Video instruction tuning with synthetic data.
\newblock \emph{arXiv preprint arXiv:2410.02713}, 2024.

\bibitem[Zhao et~al.(2024)Zhao, Gundavarapu, Yuan, Zhou, Yan, Sun, Friedman,
  Qian, Weyand, Zhao, et~al.]{zhao2024videoprism}
Long Zhao, Nitesh~B Gundavarapu, Liangzhe Yuan, Hao Zhou, Shen Yan, Jennifer~J
  Sun, Luke Friedman, Rui Qian, Tobias Weyand, Yue Zhao, et~al.
\newblock Videoprism: A foundational visual encoder for video understanding.
\newblock \emph{arXiv preprint arXiv:2402.13217}, 2024.

\bibitem[Zhou et~al.(2018)Zhou, Xu, and Corso]{zhou2018towards}
Luowei Zhou, Chenliang Xu, and Jason Corso.
\newblock Towards automatic learning of procedures from web instructional
  videos.
\newblock In \emph{Proceedings of the AAAI Conference on Artificial
  Intelligence}, volume~32, 2018.

\bibitem[Zhu et~al.(2023)Zhu, Chen, Shen, Li, and Elhoseiny]{zhu2023minigpt}
Deyao Zhu, Jun Chen, Xiaoqian Shen, Xiang Li, and Mohamed Elhoseiny.
\newblock Minigpt-4: Enhancing vision-language understanding with advanced
  large language models.
\newblock \emph{arXiv preprint arXiv:2304.10592}, 2023.

\end{thebibliography}
\bibliographystyle{iclr2025_conference}

\clearpage
\appendix

\onecolumn
{
    \hypersetup{linkcolor=black}
    \parskip=0em
    \renewcommand{\contentsname}{Contents of Appendix}
    \tableofcontents
    \addtocontents{toc}{\protect\setcounter{tocdepth}{3}}
}

\section{Dataset Preparation}
\label{sec:data-prepare}

\subsection{Details of data format}

\begin{figure}
    \centering
    \includegraphics[width=1.0\linewidth]{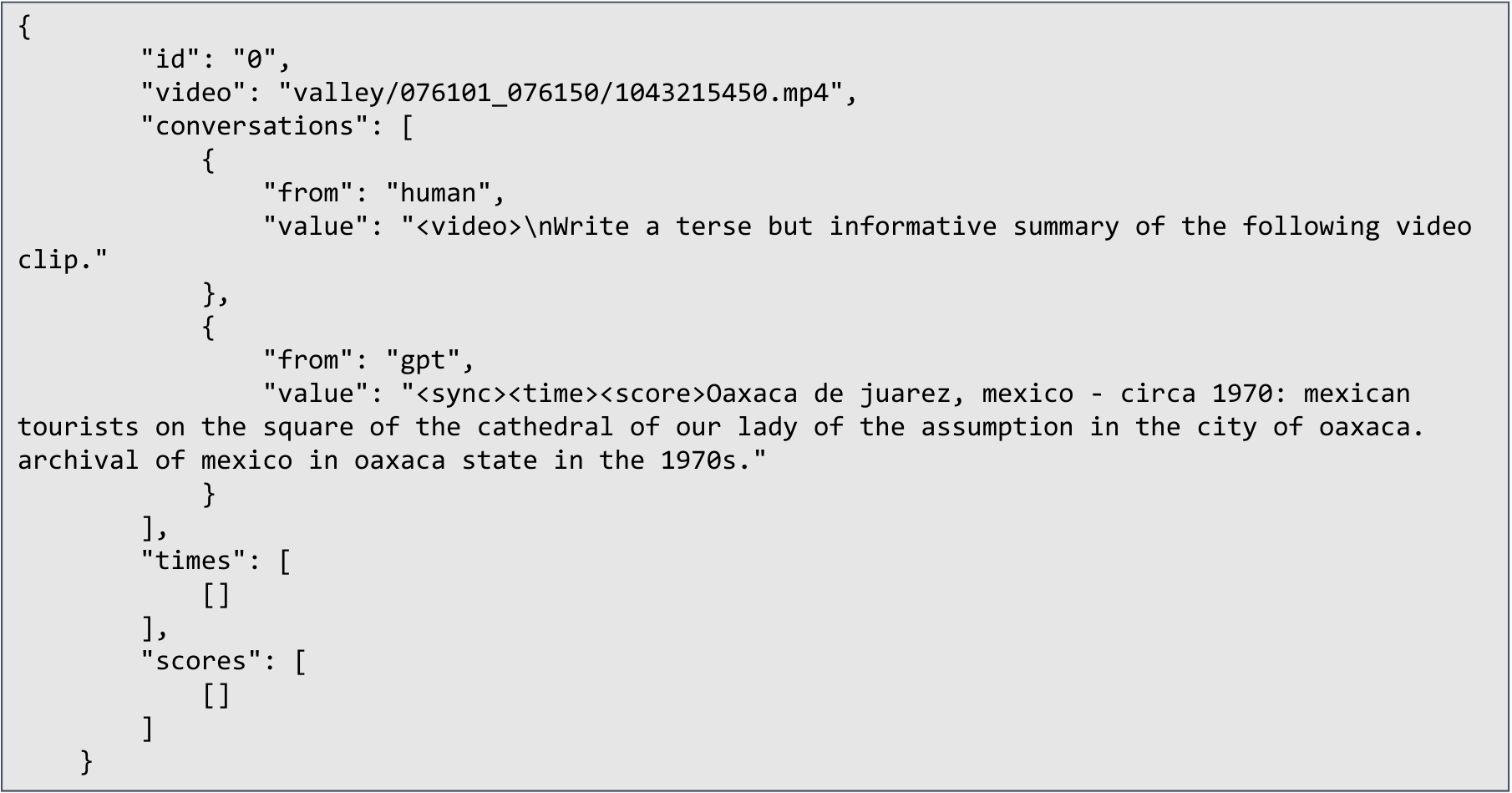}
    \caption{\small\textbf{Annotation example of video caption task.}}
    \label{fig:annotation-example-vc}
\end{figure}

\begin{figure}
    \centering
    \includegraphics[width=1.0\linewidth]{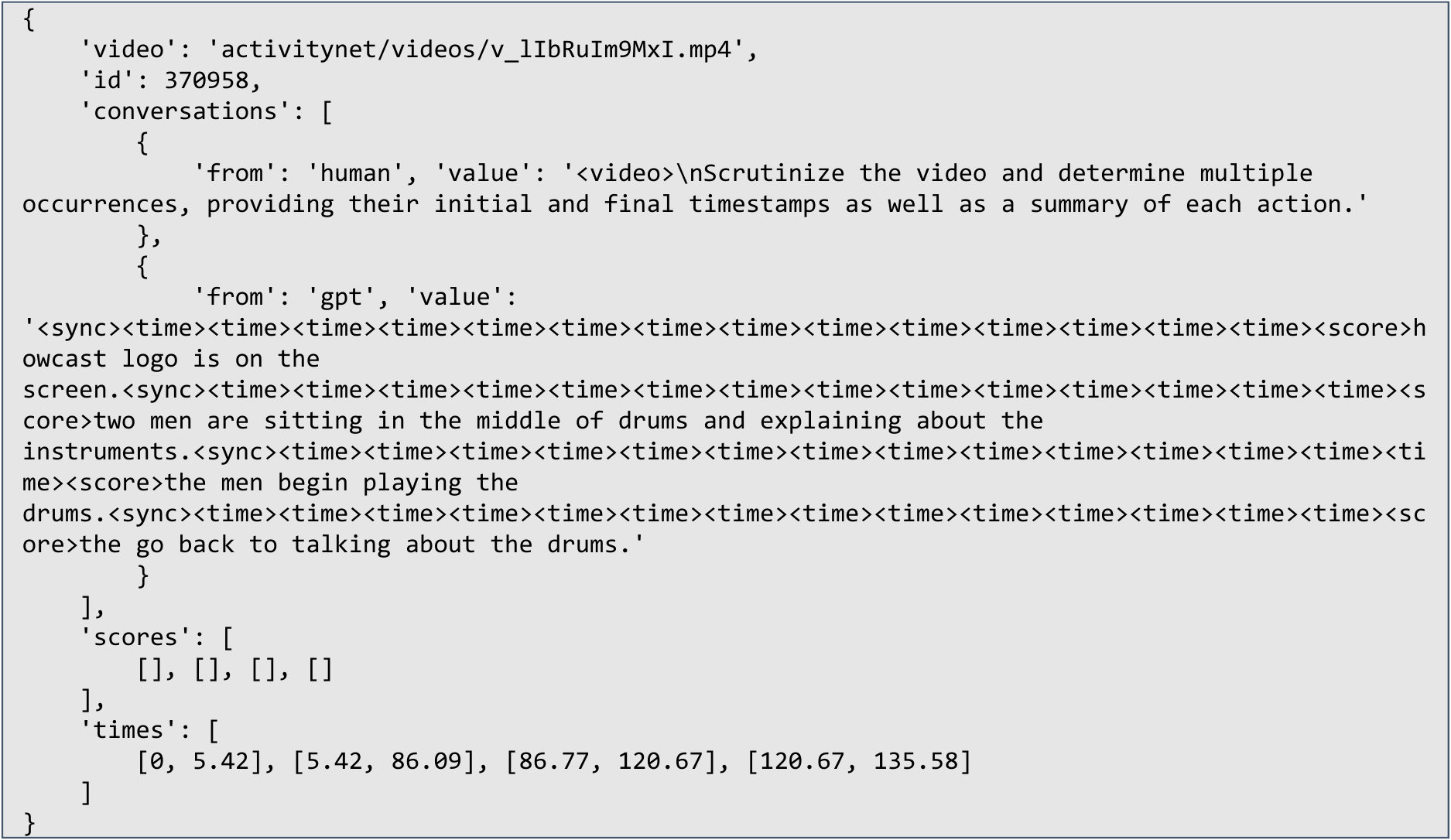}
    \caption{\small\textbf{Annotation example of dense video caption task.}}
    \label{fig:annotation-example-dvc}
\end{figure}

\begin{figure}
    \centering
    \includegraphics[width=1.0\linewidth]{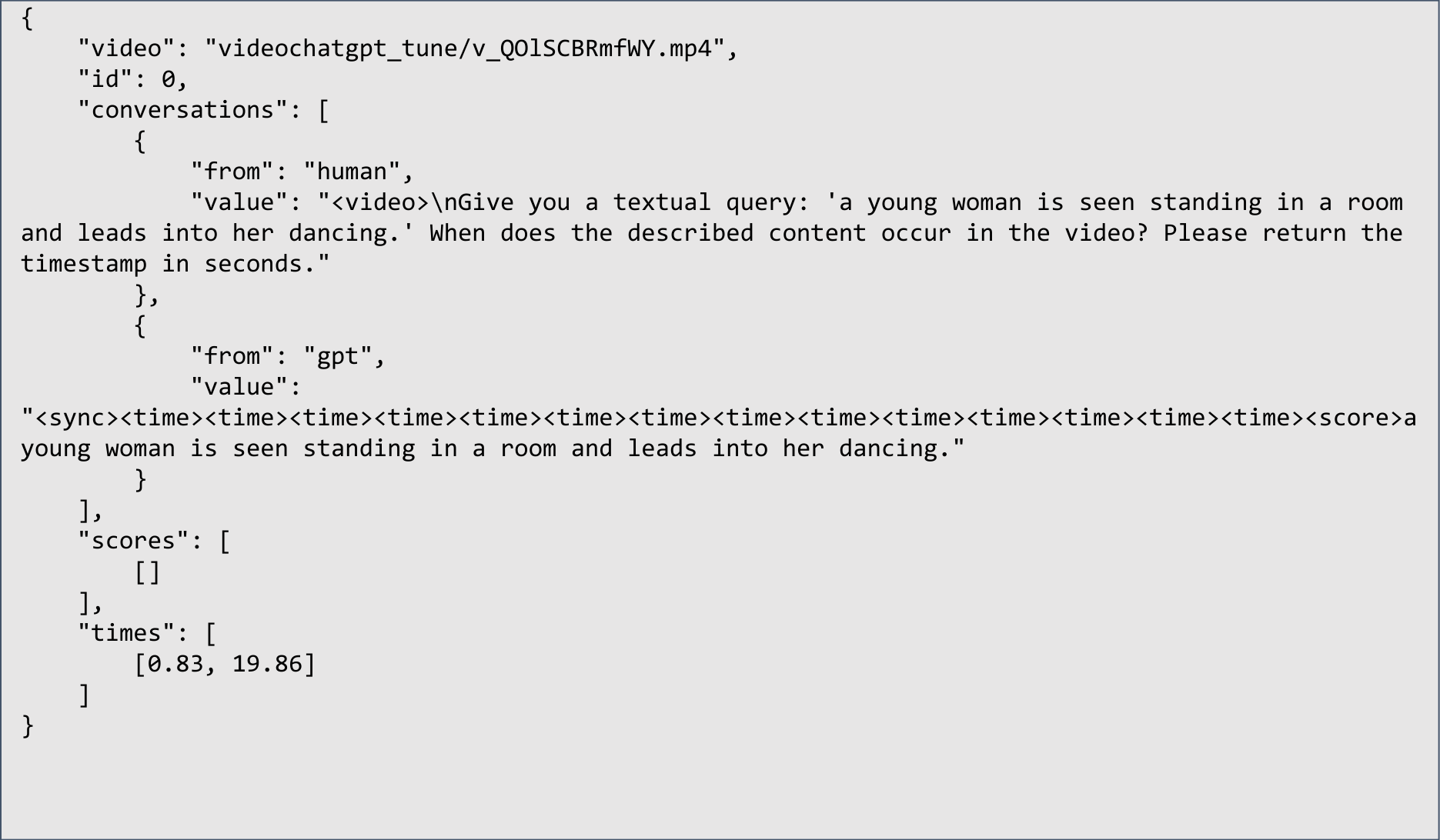}
    \caption{\small\textbf{Annotation example of moment retrieval task.}}
    \label{fig:annotation-example-mr}
\end{figure}

\begin{figure}
    \centering
    \includegraphics[width=1.0\linewidth]{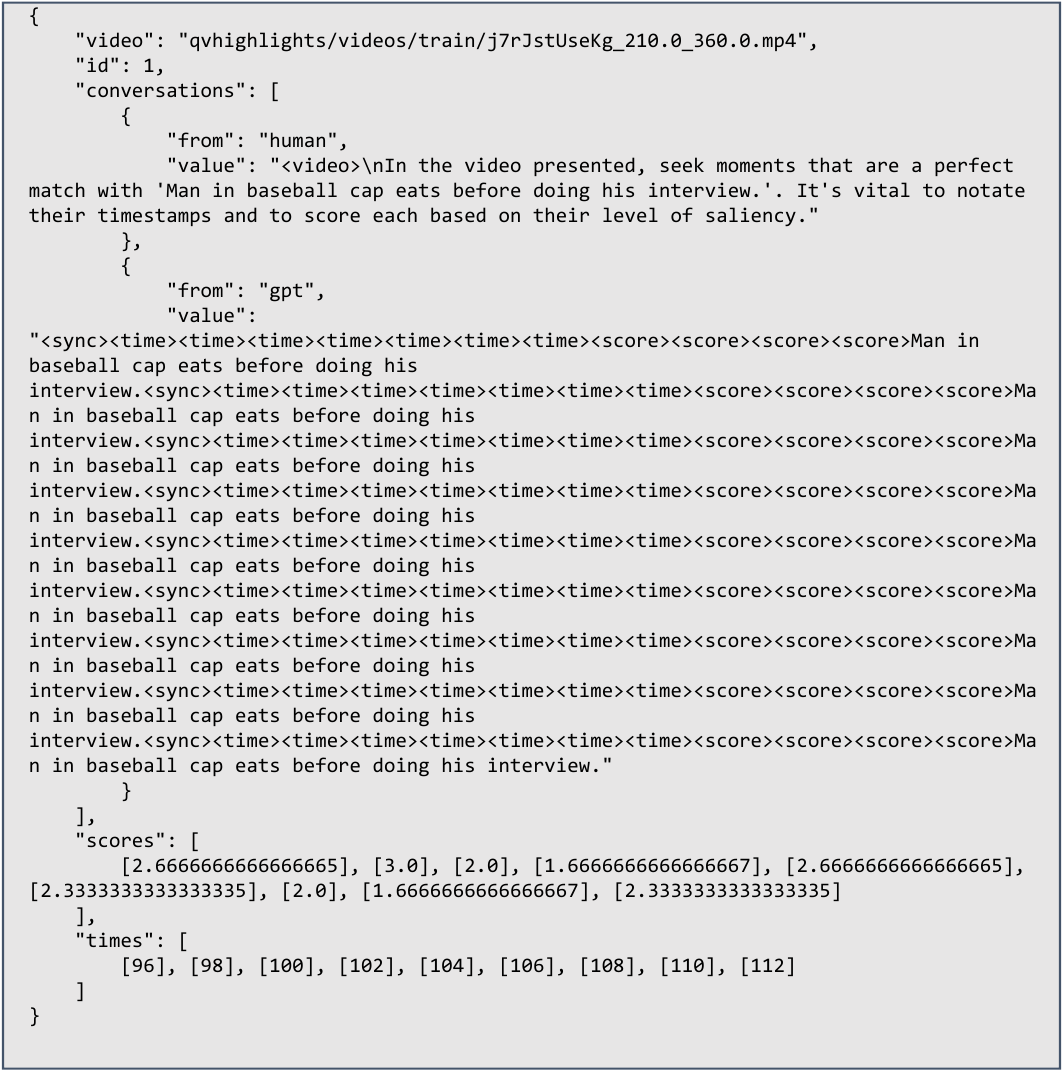}
    \caption{\small\textbf{Annotation example of video highlight detection task.}}
    \label{fig:annotation-example-vhd}
\end{figure}

\begin{figure}
    \centering
    \includegraphics[width=1.0\linewidth]{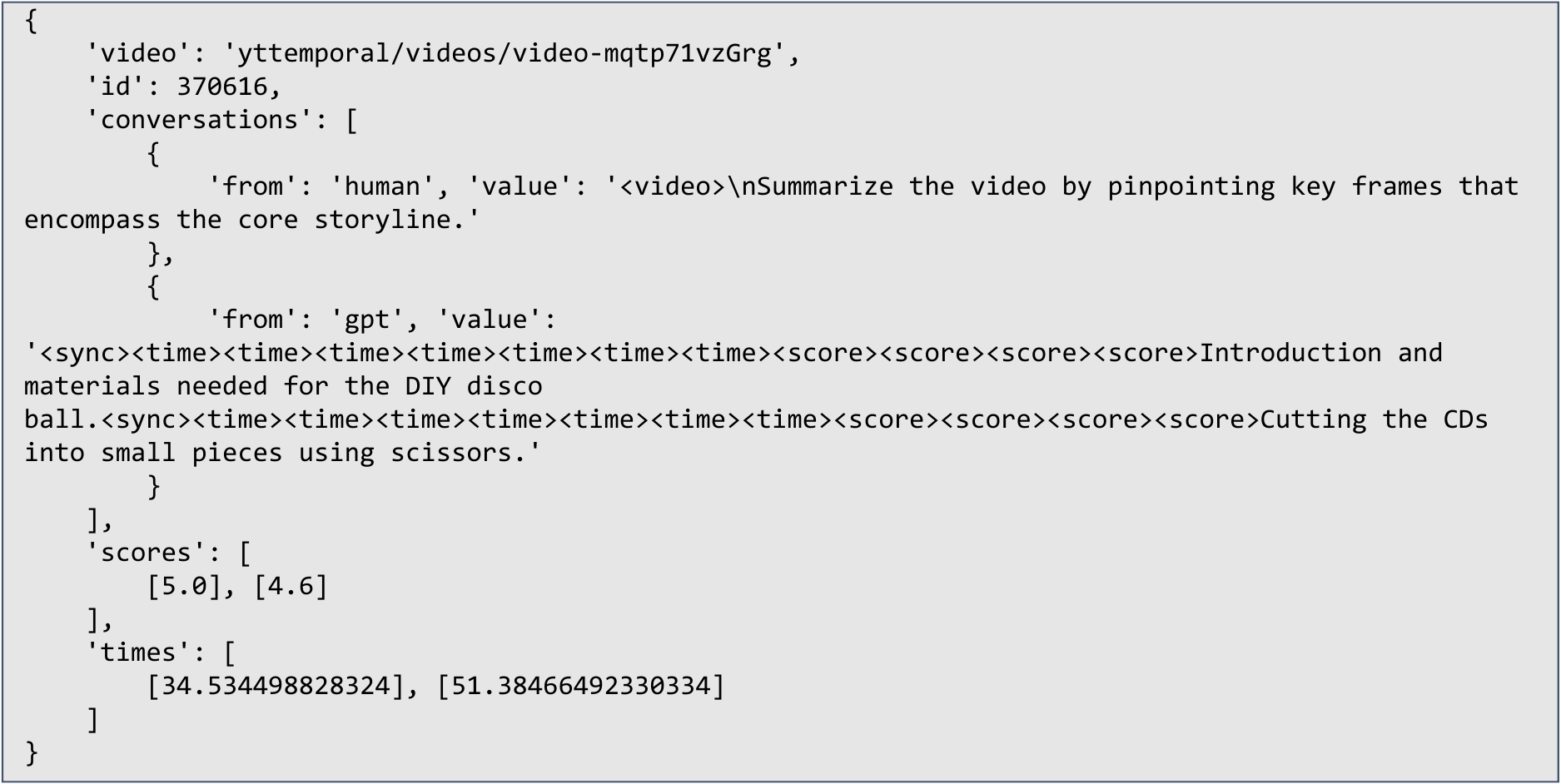}
    \caption{\small\textbf{Annotation example of video summarization task.}}
    \label{fig:annotation-example-vs}
\end{figure}

In this section, we introduce the details of the data format we used while training \alg. For the VTG datasets, in addition to ActivityNet Captions dataset and InternVid dataset, we directly use the annotation collected by VTG-IT~\citep{guo2024vtg}.
In detail, the annotations can be categories into the following four types:
\begin{itemize}[leftmargin=12pt]
    \item \textbf{General tasks (Figure~\ref{fig:annotation-example-vc}).} For general tasks such as video captioning, image captioning, and video question answering, the answer component of the data does not include timestamps or scores. Consequently, we employ a single token $\langle sync \rangle$ as a placeholder for timestamps and scores, signifying an empty response for this part of response. This kind of annotation including LLaVA\_Image~\citep{liu2024visual}, Valley~\citep{luo2023valley}, TextVR~\citep{wu2025large}, ShareGPT4Video~\citep{chen2024sharegpt4video}, VideoChatGPT~\citep{maaz2023video}, and Next-QA~\citep{xiao2021next} datasets.
    \item \textbf{Dense video caption task (Figure~\ref{fig:annotation-example-dvc}).} The Dense Video Captioning task solely comprises timestamps and textual captions responses. As a result, we use a single token $\langle sync \rangle$ as a placeholder for scores. The datasets of this task include HiREST$_{step}$~\citep{zala2023hierarchical}, COIN~\citep{tang2019coin}, ActivityNet Captions~\citep{caba2015activitynet}, VTG-IT-DVC~\citep{guo2024vtg}, and InternVid~\citep{wang2023internvid} datasets.
    \item \textbf{Moment retrieval task (Figure~\ref{fig:annotation-example-mr}).} Similiar to dense video caption task, moment retrieval solely comprises timestamps and textual captions responses. The moment retrieval task including HiREST$_{grounding}$~\citep{zala2023hierarchical}, QuerYD~\citep{oncescu2021queryd}, DiDeMo~\citep{hendricks2018localizing}, VTG-IT-MR~\citep{guo2024vtg}, and  InternVid~\citep{wang2023internvid} datasets.
    \item \textbf{Video highlight detection task (Figure~\ref{fig:annotation-example-vhd}).} For the video highlight detection task, we utilize the query as the textual response for all highlight moments. In this case, we employ the VTG-IT-VHD~\citep{guo2024vtg} dataset.
    \item \textbf{Video summarization task (Figure~\ref{fig:annotation-example-vs}).} The video summarization task employs the caption of each event as the textual response. We use the VTG-IT-VS~\citep{guo2024vtg} dataset here.
\end{itemize}

\subsection{Processing InternVid}
\label{sec:process-internvid}

\paragraph{Dense video caption data.} The dense video caption data is constructed based on the InternVid-Full annotations. For each video, we opt not to use the annotation of that video as dense video caption data if any of the following checklist items are met:
\begin{itemize}[leftmargin=12pt]        \item There exists a caption with fewer than 5 words. 
\item There exists a caption very similar to another caption within one video. We use fuzzywuzzy package here and set the threshold to 70.
    \item The number of events is less than 5 or greater than 50. \item There are special characters present, excluding letters, spaces, and dots.
\end{itemize}

\paragraph{Moment retrieval data.} We discovered that directly utilizing the InternVid-Full annotations to construct the moment retrieval data leads to suboptimal performance, likely due to the imprecise timestamp annotations. Consequently, we employ the InternVid-10M-FLT-INFO annotation to construct the moment retrieval data, which is a filtered subset of the InternVid-Full annotation provided by the authors.

\subsection{Processing VTG-IT}

In this section, we describe the processing of the VTG-IT dataset. For the dense video caption and moment retrieval tasks, we directly utilize the annotations provided by VTG-IT. However, the data for video highlight detection and video summarization tasks supplied by VTG-IT often have uniform salient scores within each video. As a result, our goal here is to enhance the quality of data for these two tasks, specifically video highlight detection and video summarization.
\begin{itemize}[leftmargin=12pt]
    \item \textbf{Video highlight detection task.} For each event in the dense video caption task, we initially divide each event into a maximum of 20 clips. Subsequently, we compute the similarity between the frames within each clip and the event captions using ViT-G/14 from EVA-CLIP~\citep{sun2023eva}. The similarity scores are then normalized to a Gaussian distribution to serve as the highlight score. In detail, the clips with scores higher than [2.275\%, 3.593\%, 5.480\%, 8.076\%, 11.507\%, 15.866\%, 21.186\%, 27.425\%, 34.458\%, 42.074\%, 50.000\%, 57.926\%, 65.542\%, 72.575\%, 78.814\%, 84.134\%, 88.493\%, 91.924\%, 94.520\%, 96.407\%, 97.725\%] of other clips will be assign scores [1.0, 1.2, 1.4, 1.6, 1.8, 2.0, 2.2, 2.4, 2.6, 2.8, 3.0, 3.2, 3.4, 3.6, 3.8, 4.0, 4.2, 4.4, 4.6, 4.8, 5.0].
    \item \textbf{Video summarization task.} The data for the video summarization task is built upon the foundation of the video highlight detection task and the dense video caption task. Specifically, for each event in the dense video caption task data, we select the clip with the highest score to serve as the summarization clip for that particular event.
\end{itemize}

\section{Experiments}

\subsection{Detailed Experimental Settings}
\label{sec:experiment-setting}

We report the detailed model architecture design and training hyper-parameters in Table~\ref{tab:training-settings}. The training takes about 5 days for stage 1 and 5 days for stage 2.

\begin{table}[!t]
    \centering
    \caption{\small\textbf{Detailed training setting and hyper-parameters.}}
    \begin{tabular}{c c c}
    \toprule
    Setting & Stage 1 & Stage 2 \\
    \midrule
    Computation & 16 ATN 910B & 16 ATN 910B \\
    Vision Encoder & openai/clip-vit-large-patch14-336 & openai/clip-vit-large-patch14-336 \\
    DeepSpeed Stage & Zero3 Offload & Zero3 Offload \\
    LLM & Mistral-7B-v0.2 & Mistral-7B-v0.2 \\
    Batch Size & 128 & 128 \\
    Num Frames & 128 & 128 \\
    Frame Sample & Uniform & Split to 128 clips then random within clip \\
    Train Epochs & 1 & 2 \\
    Learning Rate & 1e-3 & 5e-6 \\
    LR Scheduler & Cosine & Cosine \\
    Model Max Length & 4096 & 4096 \\
    \bottomrule
    \end{tabular}
    \label{tab:training-settings}
\end{table}

\subsection{Additional Experiment Results}
\label{sec:additional-experiments}

\begin{table}[!t]
    \caption{\small\textbf{Ablation studies on training data (Youcook2 and Charades-STA).}}
    \centering
    \setlength{\tabcolsep}{0.9mm}
    \fontsize{9pt}{11pt}\selectfont
    \resizebox{1.\textwidth}{!}{
    \begin{tabular}{l c c c c c c c c c c c c c}
            \toprule
            \multirow{2}{*}{Models}  &  \multicolumn{4}{c}{Youcook2}                         & \multicolumn{3}{c}{Charades-STA} \\
            \cmidrule(lr){2-5} \cmidrule(lr){6-8} 
                                        & SODA\_c & CIDEr & F1 Score & METEOR & $\text{R@1}_{\text{(IOU=0.5)}}$ & $\text{R@1}_{\text{(IOU=0.7)}}$
                                        & mIOU\\
            \midrule
            \alg (VTG-IT) & 2.1 & 7.5 & 21.4 & 2.6 & 41.2 & 20.0 & 38.9 \\
            \alg & 2.2 & 8.1 & 22.4 & 2.8 & 40.3 & 19.4 & 38.7 \\
            \alg-uni & 2.3 & 8.6 & 22.4 & 2.9 & 43.7 & 21.0 & 41.5 \\
            \bottomrule
        \end{tabular}%
        }
    \label{tab:ablation-data-youcook-charades}
\end{table}

\begin{table}[!t]
    \caption{\small\textbf{Ablation studies on training data (ActivityNet Captions).}}
    \centering
    \setlength{\tabcolsep}{0.9mm}
    \fontsize{9pt}{11pt}\selectfont
    \resizebox{1.\textwidth}{!}{
    \begin{tabular}{l c c c c c c c c c c c c c}
            \toprule
            \multirow{2}{*}{Models}  &  \multicolumn{4}{c}{Dense Video Caption}                         & \multicolumn{3}{c}{Moment Retrieval} \\
            \cmidrule(lr){2-5} \cmidrule(lr){6-8} 
                                        & SODA\_c & CIDEr & F1 Score & METEOR & $\text{R@1}_{\text{(IOU=0.5)}}$ & $\text{R@1}_{\text{(IOU=0.7)}}$
                                        & mIOU\\
            \midrule
            \alg (VTG-IT) & 5.8 & 24.7 & 38.9 & 6.0 & 19.2 & 9.3 & 25.0 \\
            \alg & 6.0 & 25.9 & 39.3 & 6.4 & 37.7 & 24.0 & 39.0 \\
            \alg-uni & 6.4 & 29.2 & 40.4 & 6.9 & 38.2 & 24.7 & 39.4 \\
            \bottomrule
        \end{tabular}%
        }
    \label{tab:ablation-data-activitynet}
\end{table}

\paragraph{Ablation studies on training data.} In Tables~\ref{tab:ablation-data-youcook-charades} and ~\ref{tab:ablation-data-activitynet}, we report the performance of \alg using different training data. 
The \alg-uni indicates incorporating additional general video understanding data from a subset of LLaVA-Video-178k~\citep{zhang2024video} (specifically the perceptiontest and YouTube parts).
The results show that (1) the performance of \alg on long videos increase while using the original \alg setting. (2) The performance of \alg slightly reduced on short videos (Charades-STA) while using the original \alg setting; (3) Although not adding more VTG data, TRACE-uni outperforms trace in both VTG tasks and general video understanding tasks.

\begin{table}[!t]
    \centering
    \caption{\small
        \textbf{Fine-tuned performance of algorithms on QVHighlights datasets.}
        We fine-tune the Algorithm on QVHighlights datasets.
        \looseness=-1
    }
        \begin{tabular}{l c c c c c c c c c}
            \toprule
            \multirow{2}{*}{Model} & \multicolumn{2}{c}{QVHighlights}                         \\
            \cmidrule(lr){2-3}
                                       & mAP                                                         & HIT@1                                                         \\
            \midrule
            TimeChat & 21.7 & 37.9 \\
            VTG-LLM & 24.1 & 41.3 \\
            \alg & \textbf{31.8} & \textbf{51.5} \\
            \bottomrule
        \end{tabular}%
    \label{tab:ft-vtg-qvhighlights}
\end{table}


\paragraph{Fine-tuned performance on QVHighlights.} In Table~\ref{tab:ft-vtg-qvhighlights}, we show the performance of \alg on QVHighlights dataset after fine-tuning. The results indicate that \alg significantly outperform other video LLMs by a large margin.

{
\paragraph{Effectiveness of \alg on general video understanding tasks.} We evaluate \alg on general video understanding tasks, and the results in Tables~\ref{tab:general-tasks-eval},~\ref{tab:etbench}, and~\ref{tab:zero-shot-vtg-appendix} show that the TRACE architecture is still capable of handling general video understanding tasks and excel in VTG tasks:
\begin{itemize}[leftmargin=12pt]
    \item Despite not being trained on extensive multi-task datasets, TRACE is still highly effective in handling general video understanding tasks.  For example, the TRACE outperform generalist video LLMs like VideoChat2, ShareGPT4Video, and ST-LLM on VideoMME benchmark.
    \item On the E.T.Bench~\citep{liu2024bench}, TRACE outperforms VideoLlama2 across all tasks; achieves performance comparable to GPT-4o on RAR, ECA, RVQ, and DVC tasks; achieves similar performance to Qwen2-VL on RVQ and GVQ tasks; and outperforms both GPT-4o and Qwen2-VL on TVG, EPM, TAL, EVS, SLC, and TEM tasks.
    \item We train TRACE-uni by incorporating additional general video understanding data from a subset of LLaVA-Video-178k~\citep{zhang2024video}(specifically the perceptiontest and YouTube parts). TRACE-uni shows both improved general video understanding and stronger VTG performance without additional VTG training data.
    \item Notably, TRACE-uni performs on par with, or even outperforms, general video LLMs that use the same LLM backbone and vision encoder (VideoLlama2) using only about 2M training data. Additionally, TRACE-uni surpasses TRACE in VTG performance across all three evaluation datasets.
\end{itemize}
\begin{table}[]
    \centering
    \caption{\small \textbf{Performance on general video understanding tasks.}}
    \begin{tabular}{c c c}
    \toprule
    Model    &  MVBench & VideoMME  \\
    \midrule
    VideoLLaMA2     & 54.6 & 46.6 \\
    \alg & 48.1 & 43.8 \\
    \alg-uni & 53.8 & 49.6 \\
    \bottomrule
    \end{tabular}
    \label{tab:general-tasks-eval}
\end{table}
}

\begin{table}[!t]
    \centering
    \caption{\small \textbf{Performance on E.T.Bench.}}
    \resizebox{1.\textwidth}{!}{
    \begin{tabular}{c c c c c c c c c c c c c c c}
    \toprule
    E.T.Bench & RAR & ECA & RVQ & TVG & EPM & TAL & EVS & VHD & DVC & DVC & SLC & SLC & TEM & GVQ \\
    Metric & Acc & Acc & Acc & F1 & F1 & F1 & F1 & F1 & F1 & Sim & F1 & Sim & Rec & Acc \\
    \midrule
    VideoLLama2 (7B) & 28.8 & 27.4 & 28.0 & 0.1 & 0.0 & 0.0 & 0.0 & 1.5 & 0.6 & 14.5 & 0.0 & 15.2 & 0.0 & - \\
    Qwen2-VL (7B) & 39.4 & 34.8 & 42.2 & 3.9 & 0.1 & 0.3 & 0.4 & 20.6 & 0.0 & 0.0 & 0.0 & 0.0 & 6.6 & 55.9 \\
    GPT-4o & 27.8 & 27.3 & 57.7 & 40.4 & 4.5 & 20.0 & 17.6 & 56.9 & 46.9 & 22.3 & 23.1 & 14.9 & 13.6 & - \\
    TRACE (7B) & 29.4 & 28.8 & 42.6 & 46.8 & 12.3 & 21.6 & 26.6 & 45.2 & 45.7 & 24.0 & 27.3 & 17.7 & 17.8 & 52.4 \\
    \bottomrule
    \end{tabular}
    }
    \label{tab:etbench}
\end{table}

\begin{table*}[!t]
    \centering
    \caption{\small
        \textbf{Zero-shot performance of algorithms over various tasks.}
        We evaluated the performance of \alg using the Youcook2, Charades-STA, and QVHighlights datasets. 
        We highlight the best results for each block using \textbf{bold font}. The Valley, VideoChat-Embed, and Video-LLaMA results are elaborated from previous studies~\citep{ren2023timechat,huang2023vtimellm,qian2024momentor}. 
        The results with transparent text indicates unfair comparison (13B).
        We train TRACE-uni by incorporating additional general video understanding data from a subset of LLaVA-Video-178k~\citep{zhang2024video}.
        %
    }
    \setlength{\tabcolsep}{0.9mm}
    \fontsize{9pt}{11pt}\selectfont
    \resizebox{1.\textwidth}{!}{
        \begin{tabular}{l c c c c c c c c c}
            \toprule
            \multirow{2}{*}{Model} & \multicolumn{3}{c}{Youcook2}                         & \multicolumn{2}{c}{Charades-STA}                      & \multicolumn{2}{c}{QVHighlights}                                                                                                                                                                                                                                    \\
            \cmidrule(lr){2-4} \cmidrule(lr){5-6} \cmidrule(lr){7-8}
                                       & SODA\_c                                                          & CIDEr                                                         & F1 Score                                                          & $\text{R@1}_{\text{(IOU=0.5)}}$                                                         & $\text{R@1}_{\text{(IOU=0.7)}}$                                                                   & mAP & HIT@1                                                                 \\
            \midrule
            \textit{\textbf{\small Traditional Video LLMs}} \\
            Valley (7B) & 0.1 & 0.0 & 1.5 & 4.7 & 1.6 & 10.9 & 15.2                   \\
            VideoChat (7B) & 0.2 & 0.6 & 3.4 & 3.2 & 1.4 & 13.1 & 18.1 \\
            Video-LLaMA (7B) & 0.0 & 0.0 & 0.1 & 2.7 & 1.2 & 11.3 & 15.6 \\
            \midrule
            \textit{\textbf{\small Temporal Grounding Video LLMs}} \\
            TimeChat (7B) & 1.2 & 3.4 & 12.6 & 32.2 & 13.4 & 14.5 & 23.9\\
            VTimeLLM (7B) &  &  &  & 27.5 & 11.4 &  & \\
            {\texttransparent{0.5}{VTimeLLM (13B)}} &  &  &  & \texttransparent{0.5}{34.3} & \texttransparent{0.5}{14.7} & & \\
            Momentor (7B) &  &  &  & 26.6 & 11.6 & 7.6 &  \\
            HawkEye (7B) &  &  &  & 31.4 & 14.5 &  & \\
            VTG-LLM (7B) & 1.5 & 5.0 & 17.5 & 33.8 & 15.7 & 16.5 & 33.5 \\
            \midrule
            \alg (7B) & 2.2 & 8.1 & \textbf{22.4} & 40.3 & 19.4 & 26.8 & 42.7\\
            \alg-uni (7B) & \textbf{2.3} & \textbf{8.6} & \textbf{22.4} & \textbf{43.7} & \textbf{21.0} & \textbf{27.5} & \textbf{43.9} \\
            \bottomrule
        \end{tabular}%
    }
    \label{tab:zero-shot-vtg-appendix}
\end{table*}

\begin{table}[!t]
    \centering
    \caption{\small \textbf{Performance of \alg with different number of slots for each token on Youcook2 dataset.} We only use VTG-IT for training in Stage 2, and sample 64 frames for each video.}
    \begin{tabular}{c c c c c c}
    \toprule
    Frame Num & Slot Num per Frame & SODA\_c & CIDEr & F1 Score \\
    \midrule
    64 & 8 & 1.9 & 6.9 & 21.4 \\
    64 & 16 & 2.1 & 7.3 & 22.1 \\
    \bottomrule
    \end{tabular}
    \label{tab:slot-youcook2}
\end{table}

\begin{table}[!t]
    \centering
    \caption{\small \textbf{Performance of \alg with different number of slots for each token on Youcook2 dataset.} We only use VTG-IT for training in Stage 2, and sample 64 frames for each video.}
    \begin{tabular}{c c c c c c}
    \toprule
    Frame Num & Slot Num per Frame & R@1$_{IOU=0.5}$ & R@1$_{IOU=0.7}$ \\
    \midrule
    64 & 8 & 37.0 & 17.0 \\
    64 & 16 & 41.9 & 20.1 \\
    \bottomrule
    \end{tabular}
    \label{tab:slot-charades}
\end{table}

\paragraph{Performance of \alg with different number of slots per frame.}  In Tables~\ref{tab:slot-youcook2} and~\ref{tab:slot-charades}, we adopted the same settings as in Table~\ref{tab:ablation}, using VTG-IT only in Stage 2 and sampling 64 frames. Our findings are as follows: Increasing the number of slots per token significantly enhances TRACE's performance. Therefore, if computational or efficiency constraints are not a concern, we recommend using a larger number of slots per frame.


\subsection{Case Studies}
\label{sec:case-study}

\begin{figure}
    \centering
    \includegraphics[width=1.\linewidth]{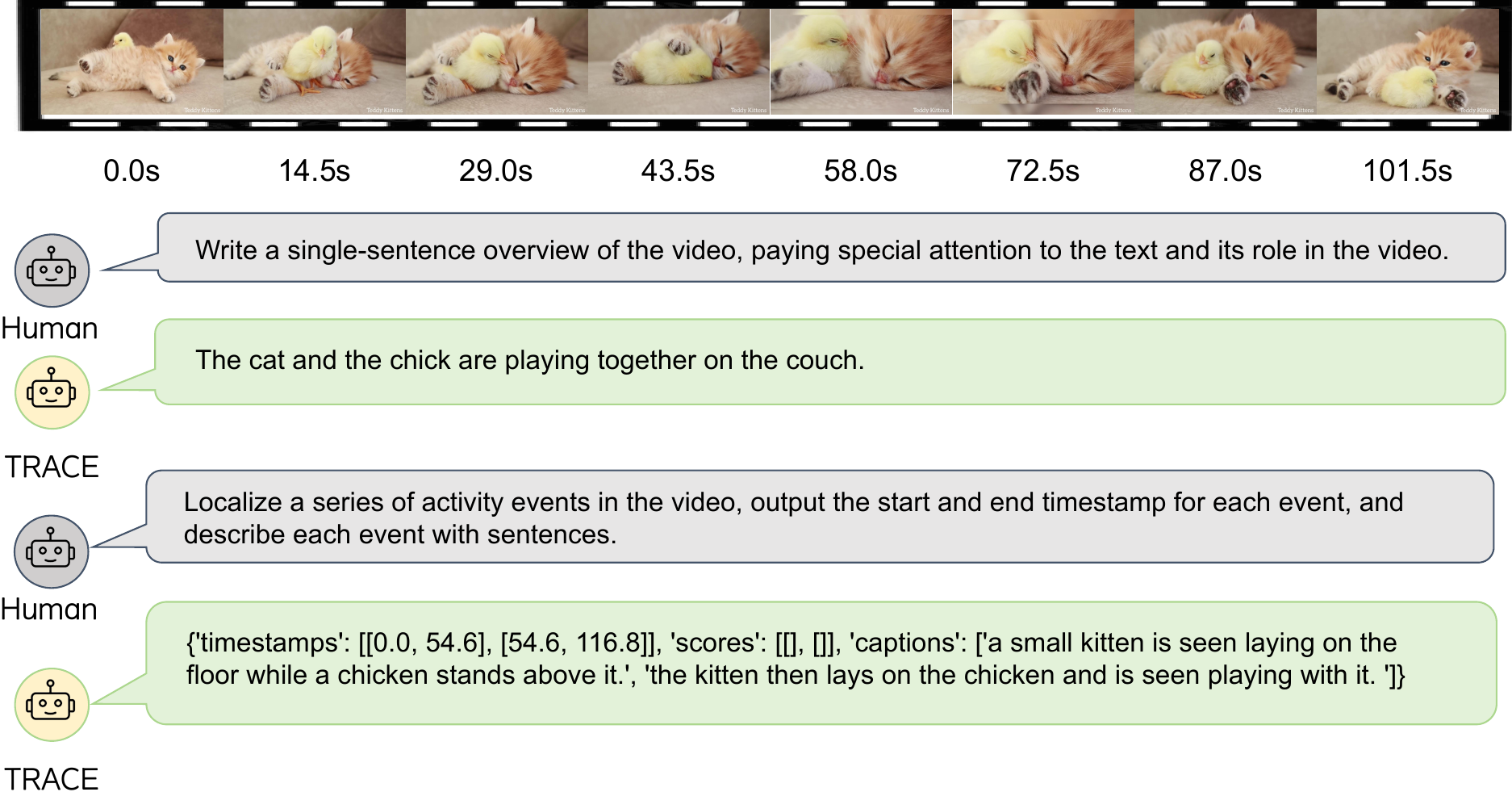}
    \caption{\small \textbf{Case study of \alg.}}
    \label{fig:case-study}
\end{figure}

We present the case studies of \alg in Figure~\ref{fig:case-study}. The results demonstrate that \alg can accurately identify the events within the given video and is also proficient in performing traditional video captioning tasks.

\section{{Discussion on Causal Event Modeling}}
\label{sec:discussion}

{
In this section, we provide a deeper analysis of the causal modeling approach presented in this paper. We also offer a comprehensive discussion and comparison of the following three key concepts:
\begin{itemize}[leftmargin=12pt]
    \item \textbf{Causal Language Modeling.} This is a well-established approach for decoder-only large language models (LLMs) and have become basis of current video LLMs~\citep{li2023videochat,ren2023timechat,huang2023vtimellm}
    \item \textbf{Causal Event Modeling}. In contrast, \alg introduces causal event modeling, which structures the responses of video LLMs by event triplets, providing a novel framework for understanding video content.
    \item \textbf{Complete Causal Relationship Modeling/Discovery.} Traditional video understanding models~\citep{liang2022visual,parmar2024causalchaos,chen2024mecd,du2024towards,yi2019clevrer,girdhar2019cater,li2020causal,jin2022complex} focus extensively on discovering and analyzing the complex relationships between events within videos, offering a more comprehensive understanding of video content.
\end{itemize}
We would like to first discuss how causal event modeling excels in video understanding compared to causal language modeling. Next, we will explore the benefits, limitations, and potential integration of \alg compared to complete causal relationship modeling/discovery methods.
\subsection{Causal Language Modeling VS. Causal Event Modeling.} We will first provide a clear comparison between causal language modeling and causal event modeling, highlighting their similarities and differences. Then, we will delve into the benefits of using causal event modeling in more detail.
}

{
\textbf{Similarities.} Both causal language modeling and causal event modeling rely on decoder-only LLMs, which use video visual content ($\mF$) and instructions ($\mI$) as inputs. The models then generate responses based on the observations from both $\mI$ and $\mF$.
}

{
\textbf{Differences.} The key differences come from the output formats.
}

{
For causal language modeling, the response can be seen as a whole text content, i.e.,
\begin{align}
    p(\mY|\mF, \mI) &= \prod_{y} p(y | y_{i} < y, \mF, \mI) \, .
\end{align}
Here, timestamps, scores, and text are all represented by text tokens, and ordered following the natrual language structure.
}

{
For causal event modeling, the responses are formatted to series of event triplets $e_i = (t_i, s_i, c_i)$, i.e.,
\begin{align}
    p(\mR|\mF, \mI) = \prod_{e_{i}} p(e_{i} | e_{1:i}, \mF, \mI) = \prod_{e_{i}} p( t_i | e_{1:i}, \mF, \mI) p( s_i | t_i, e_{1:i}, \mF, \mI) p( c_i | s_i, t_i, e_{1:i}, \mF, \mI) \, .
\end{align}
}

{
\textbf{Key Improvements.} We can find that the key improvement of causal event modeling over causal language modeling comes from the format of model responses. 
\begin{itemize}[leftmargin=12pt]
    \item \textbf{Clear inter- and intra- event triplets correlation.} The inter-event relationships are modeled through the next event prediction formulation. For intra-event relationships, we observe that a causal connection exists between $ t_i \to s_i $, and between $ (t_i, s_i) \to c_i $. In this context, $ t_i $ and $s_i$ can also be viewed as a kind of thought chain when generating $ c_i $.
    \item \textbf{Enable independent timestamps, scores, and text modeling.} We can separate the modeling of timestamps $t_i$, scores $s_i$, and text $c_i$, and independently model each component using different encoders and decoders. As directly adding new tokens to the text tokenizer may significantly disrupting the pretrained LLMs~\citep{guo2024vtg}, such a decomposition helps to eliminate this issue, and the \alg architecture retain the capacity on handling general video understanding tasks, as evaluated in Table~\ref{tab:general-tasks-eval}.
\end{itemize}
}

{
\subsection{Causal Event Modeling VS. Complete Causal Relationship Modeling/Discovey}
In this section, we will discuss the difference between causal event modeling and studies~\citep{liang2022visual,parmar2024causalchaos,chen2024mecd,du2024towards,yi2019clevrer,girdhar2019cater,li2020causal,jin2022complex} focused on discovering or modeling complete causal relationships. Additionally, we will explore potential future work on integrating these two approaches.
\paragraph{Discussion on studies that discovering or modeling complete causal relationships.}
Based on the existing studies in this area, we classify the research into three main groups, which we will discuss in detail:
\begin{itemize}[leftmargin=12pt]
    \item \textbf{Building complete causal relationships to solve the video understanding problems.}  VAR~\citep{liang2022visual} uses an encoder-decoder architecture that first encodes video frames into event representations, then decodes each event into text captions. Similarly, \citet{jin2022complex} constructs a learnable Markov Logic Network for action reasoning. Both of these approaches model complex causal relationships between events before generating answers, thereby enhancing the reasoning capacity of the models.
    \item \textbf{Introducing Benchmark Datasets for Complex Causal Reasoning.} Several studies have introduced benchmark datasets to evaluate models' ability to perform causal reasoning~\citep{parmar2024causalchaos, yi2019clevrer, girdhar2019cater, du2024towards}.
    \item \textbf{Discovering Causality from Videos.} Some research aims to automatically build causality graphs to uncover the relationships between events, providing a more comprehensive understanding of video content~\citep{chen2024mecd, li2020causal}.
\end{itemize}
\paragraph{Evaluation results of \alg on causality reasoning benchmarks.} In Table~\ref{tab:causal-bench}, we present the performance of \alg on the Event-Bench~\citep{du2024towards}.The results show that, despite being trained on less causal reasoning data (NextQA) compared to other video LLMs \citep{li2023mvbench}, \alg outperforms open-source video LLMsand achieves performance comparable to GPT-4o on event description, contextural reasoning, and episodic reasoning tasks.
\paragraph{Potential future improvements to TRACE through integration with causality discovery models.}
Current video LLMs typically generate answers directly from text prompts and visual frames without explicitly modeling the causal relationships between events. While \alg improves upon this limitation by representing model responses as event triplets, it still relies on previously generated events to produce subsequent triplets, due to the inherent architecture of decoder-only LLM backbones. To address this issue, further improvements could be made by integrating causality discovery methods:
\begin{itemize}[leftmargin=12pt]
    \item \textbf{Utilizing the outputs of causality discovery models as inputs for video LLMs.}
    For instance, we can encode the causality graph produced by \citet{chen2024mecd, li2020causal} as part of the model's inputs. For \alg, this involves modifying $p(\mR|\mF, \mI)$ to $p(\mR|\mF, \mI, \mC)$, where $\mC$ represents the generated causality graph. By incorporating the causality graph as an additional input, the model would have access to richer context, enabling it to generate more accurate responses.
    \item \textbf{Utilizing the outputs of causality discovery models to construct Chain-of-Thought examples.} We can also guide video LLMs to first generate causality graphs before answering questions, similar to the approach used by \citet{jin2022complex}. For \alg, we can modify $p(\mR|\mF, \mI)$ to $p(\mC, \mR|\mF, \mI) = p(\mC|\mF, \mI)p(\mR|\mC, \mF, \mI)$.
    \item \textbf{Utilizing the outputs of causality discovery models to modify the attention masks of visual inputs.} Currently, the attention masks for visual inputs are typically designed the same as text tokens, i.e., using causal attention masks. However, by incorporating the causality graph, we can enhance the attention masks by masking out visual tokens that are not causally related to the current reasoning task, thereby improving model focus on relevant events.
\end{itemize}
}


\begin{table}[!t]
    \caption{ \small\textbf{Evaluation results on Event-Bench.}}
    \centering
    \setlength{\tabcolsep}{0.9mm}
    \fontsize{9pt}{11pt}\selectfont
    \resizebox{1.\textwidth}{!}{
    \begin{tabular}{l c c c c c c c c c c c c c}
            \toprule
            \multirow{2}{*}{Models}  &  \multicolumn{1}{c}{Atomic}                         & \multicolumn{3}{c}{Composite} & \multicolumn{4}{c}{Overall} & \multirow{2}{*}{Avg} \\
            \cmidrule(lr){2-2} \cmidrule(lr){3-5} \cmidrule(lr){6-9} 
                                        & Event Description & Temporal Reasoning &	Causal Reasoning &	Avg & Counter Reasoning &	Contextual Reasoning &	Episodic Reasoning &	Avg. &
\\
            \midrule
            Video-LLaVA (7B) &	12.82 &	5.50 &	0.00 &	2.75 &	6.17 &	2.78 &	7.20 &	5.05 &	5.87 \\
            VideoChat2 (7B) &	33.76 &	37.75 &	47.75 &	42.75 &	16.74 &	15.70 &	14.67 &	15.62 &	29.41 \\
            ST-LLM (7B) & 47.22 & 48.75 & 59.50 & 54.13 & 9.69 & 25.32 & 16.67 & 18.66 & 37.71 \\
             VIM (7B) & 48.08 &	51.25 &	61.25 &	56.25 &	22.91 & 32.66 &	18.67 &	25.71 	&41.64 \\
             Gemini-1.5-Pro & 48.50 & 47.50 & 41.75 & 44.63 & 52.86 & 32.15 & 38.67 & 39.37 & 43.24 \\
             GPT-4o &         	54.27 &	56.75 	& 58.25 &	57.50 &	63.44 &	50.13 &	37.33 
 &	49.24 &	53.33 \\
            \alg (7B) & 55.56 & 49.25 &	54.50 &	51.88 & 	52.42 &	46.08 &	43.00 &	46.64 &	50.46 \\

            \bottomrule
        \end{tabular}%
        }
    \label{tab:causal-bench}
\end{table}

\end{document}